\documentclass[10pt, a4paper]{article}

\usepackage[]{lrec2026} 
\usepackage{paralist} 
\usepackage{threeparttable}
\usepackage{booktabs} 
\usepackage{multirow} 
\usepackage{amsmath}
\usepackage{amssymb} 
\usepackage{multibib}
\usepackage{array} 
\usepackage{float}
\usepackage{newunicodechar}
\newunicodechar{−}{\textminus}

\usepackage{graphicx} 
\usepackage{subcaption}  
\usepackage{needspace}
\usepackage{tabularx} 
\usepackage[margin=1in]{geometry}

\newcites{LR}{Language Resource References}

\title{BURMESE-SAN: Burmese NLP Benchmark \\for Evaluating Large Language Models}

\name{
    Thura Aung$^{1,*}$\thanks{* Work conducted during Research Internship at AI Singapore, National University of Singapore.},
    Jann Railey Montalan$^{2,3}$,
    Jian Gang Ngui$^{2,3}$,
    Peerat Limkonchotiwat$^{2,3}$
}

\address{
    $^{1}$King Mongkut's Institute of Technology Ladkrabang \\
    $^{2}$AI Singapore $^{3}$ National University of Singapore \\
    \texttt{66011606@kmitl.ac.th} \\
    \texttt{\{railey, jiangangngui, peerat.limkonchotiwat\}@aisingapore.org}
}

\abstract{
We introduce \emph{BURMESE-SAN}, the first holistic benchmark that systematically evaluates large language models (LLMs) for Burmese across three core NLP competencies: understanding (NLU), reasoning (NLR), and generation (NLG).
BURMESE-SAN consolidates seven subtasks spanning these competencies, including Question Answering, Sentiment Analysis, Toxicity Detection, Causal Reasoning, Natural Language Inference, Abstractive Summarization, and Machine Translation, several of which were previously unavailable for Burmese.
The benchmark is constructed through a rigorous native-speaker-driven process to ensure linguistic naturalness, fluency, and cultural authenticity while minimizing translation-induced artifacts.
We conduct a large-scale evaluation of both open-weight and commercial LLMs to examine challenges in Burmese modeling arising from limited pretraining coverage, rich morphology, and syntactic variation.
Our results show that Burmese performance depends more on architectural design, language representation, and instruction tuning than on model scale alone. In particular, southeast asia regional fine-tuning and newer model generations yield substantial gains.
Finally, we release \emph{BURMESE-SAN} as a public leaderboard to support systematic evaluation and sustained progress in Burmese and other low-resource languages. \url{https://leaderboard.sea-lion.ai/detailed/MY}.
 \\ \newline  }

\begin{document}

\maketitleabstract


\section{Introduction}
\label{sec:intro}
Recent advances in Large Language Models (LLM) development have improved the overall capabilities of Natural Language Processing (NLP). 
With a large amount of data and high computing power, LLMs have seen widespread adoption due to their ability to provide scalable, practical, and diverse capabilities \cite{guo2023evaluating,hadi2023survey}.
In particular, studies~\cite{liang2023holistic, QIN2025101118, 10.1145/3641289} demonstrate the effectiveness of LLMs across various languages and tasks, utilizing their benchmarks to evaluate and compare model performance.  

HELM~\cite{liang2023holistic} proposed formulating a holistic benchmark to evaluate the robustness of LLMs for a diversity of tasks in English.
SEACrowd~\cite{lovenia-etal-2024-seacrowd} and SEA-HELM~\cite{susanto2025seahelmsoutheastasianholistic}, designed as holistic benchmarks, evaluate LLMs' linguistic capabilities in several Southeast Asian (SEA) languages, and both use human-crafted and machine-generated corpora.
The fact that some LLMs can perform well on these benchmarks demonstrates that LLMs can generalize from English to SEA languages.

However, most existing LLM benchmarks focus on English, with relatively few evaluating non-English languages~\cite{son2024mm}. 
Furthermore, even in multilingual benchmarks~\cite{liang2023holistic, susanto2025seahelmsoutheastasianholistic}, Burmese is not included because the necessary datasets are not yet available. 
Despite being the official language of Myanmar and having a total number of speakers of over 43-45 million \cite{ethnologue2019mya}, Burmese remains an under-resourced language for both models and benchmarks~\cite{dou2025sailor2sailingsoutheastasia}.

\begin{figure*}[h!]
    \centering
    \includegraphics[width=0.9\linewidth]{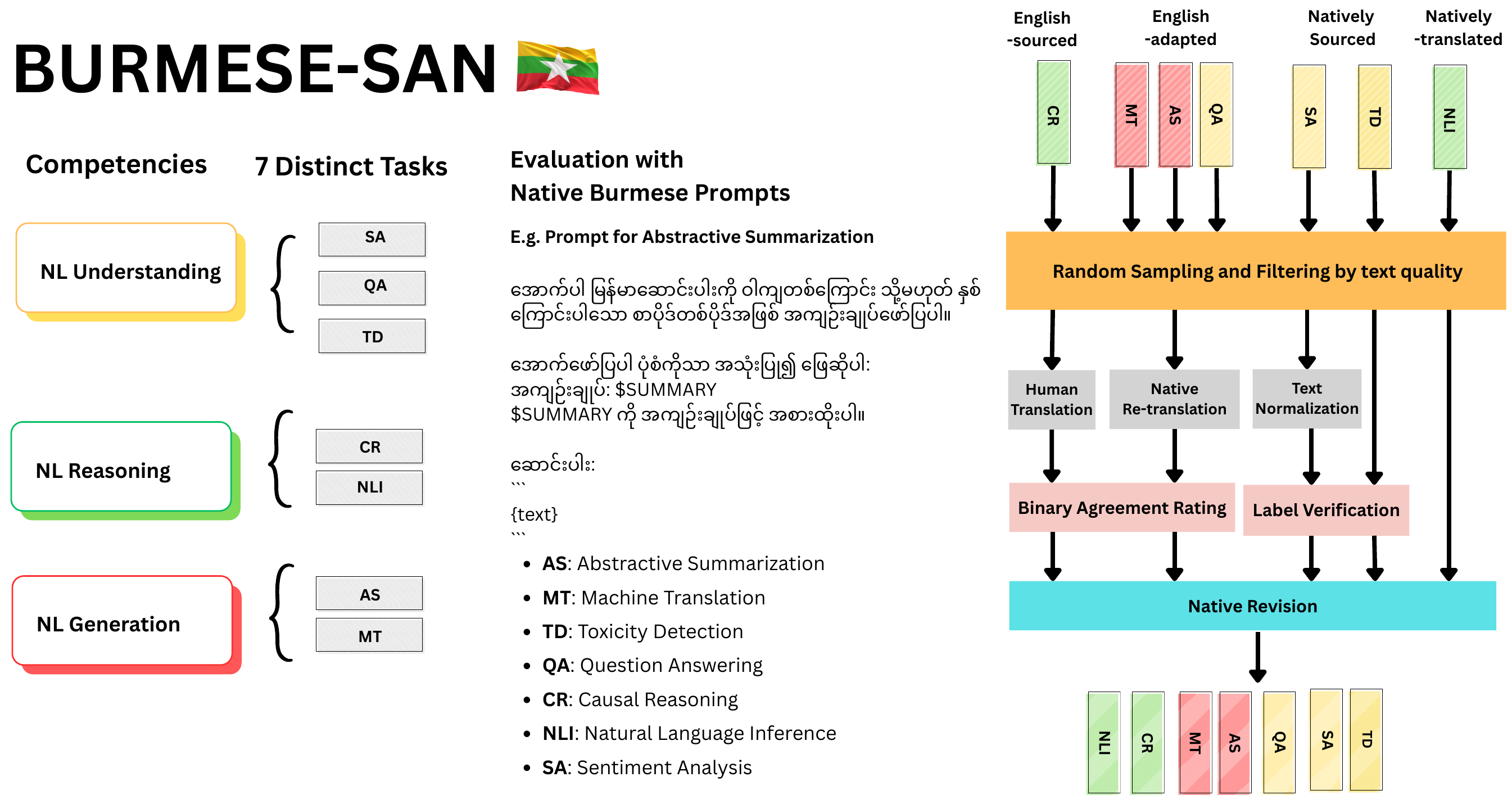}
    \vspace{-3mm}
    \caption{\textbf{\emph{BURMESE-SAN} Benchmark (Left) and Dataset Curation Process for the benchmark (Right).} \emph{BURMESE-SAN} is a benchmark that holistically evaluates LLM performance across a wide range of Burmese language tasks. The evaluation is based on native Burmese text, with prompts written in formal Burmese to ensure clarity and grammatical correctness.}
    \label{fig:task_summary}
\end{figure*}

Burmese stands out among the major SEA languages for its extensive system of case marking and its rich morphological structure \cite{jenny2021msea}.
It is a tonal language with an analytic grammatical structure \cite{okell2023burmese}. Burmese features four distinct tones, with each syllable carrying a tone that plays a crucial role in distinguishing meaning between otherwise identical syllables.

The language follows a subject-object-verb (SOV) word order and utilizes a complex system of honorifics and politeness markers that reflect social hierarchy and relationships \cite{myanmar2013burmesegrammar, myanmarspokengrammar}.
Burmese is also a diglossic language, with formal and informal varieties that can differ significantly in vocabulary and structure, adding another layer of complexity for language modeling and evaluation.
Therefore, due to the challenge of language and the availability of Burmese corpora, we need to dedicate more resources to creating a high-quality benchmark to identify any gaps and challenges in current LLMs.

To tackle these challenges, we propose \emph{BURMESE-SAN}\footnote{The word \textit{San} means \textit{Standard} in Burmese.}, \emph{the first holistic Burmese Benchmark} for evaluating LLMs across \emph{seven} NLP tasks covering natural language understanding, reasoning, and generation. 
In particular, as shown in Figure~\ref{fig:task_summary}, we have seven subtasks: sentiment analysis (SA), toxicity detection (TD), question answering (QA), causal reasoning (CR), natural language inference (NLI), abstractive summarization (AS), and machine translation (MT), with a total of 3,920 samples. 
%
%
\emph{BURMESE-SAN} was meticulously built in collaboration with native speakers, who have been involved at every stage of the pipeline  -  from task selection, prompt translation, and data validation to final quality assurance. 
This high-effort, \emph{human-centered} approach ensures linguistic authenticity, cultural relevance, and evaluative rigor, setting a new standard for Burmese benchmarks. 

\noindent \textbf{Proposed Studies.}
Based on \emph{BURMESE-SAN}, we investigate the following research questions:

\begin{compactenum}[\bf RQ1]
    \item \textbf{Comparison of Commercial and Open-Source Models}
    \emph{How do commercial LLMs compare with open-weight models on Burmese language tasks?}

    \item \textbf{Effect of Model Scale}
    \emph{Does increasing model size lead to improved performance on Burmese NLP tasks?}

    \item \textbf{Effect of Southeast Asian (SEA) Fine-Tuning}
    \emph{Does fine-tuning on Southeast Asian languages improve model performance on Burmese?}

    \item \textbf{Effect of Model Quantization}
    \emph{How does model quantization affect performance compared to full-precision models?}

    \item \textbf{Temporal Progress in Burmese Language Capability}
    \emph{Have LLMs demonstrated consistent improvements in Burmese performance across model generations?}
\end{compactenum}

\noindent \textbf{Contributions.}
We summarize the contribution of our work as follows:
\begin{compactitem}
    \item We propose \emph{BURMESE-SAN}. A holistic Burmese benchmark for evaluating LLMs across seven NLP tasks. Our benchmark is formulated and edited by humans, resulting in high-quality samples.
    \item  We present a reproducible dataset development process, covering sourcing, task design, and quality checks. \url{https://github.com/aisingapore/SEA-HELM}.
    \item We conduct extensive experiments on Instruction-Tuned, and Reasoning LLMs of different sizes, revealing systematic gaps in Burmese understanding, reasoning, and generation. The results show that SEA-tuned models can substantially improve performance.
\end{compactitem}


\section{Related Works}
\subsection{Burmese NLP}
Myanmar is a linguistically diverse country with around 100 ethnic languages and dialects, belonging to four major language families: Sino-Tibetan, Austro-Asiatic, Tai–Kadai, and Indo-European \cite{ethnologue_myanmar_2024}. Although Standard Burmese is the official language, several dialects are spoken, including Beik, Dawei, and Rakhine \cite{mydialects}.

Burmese (or Bamar), also known as Standard Burmese, is the native language of the Bamar majority and is also spoken by related groups such as the Mon. It belongs to the Sino-Tibetan (specifically Tibeto-Burman) language family. Burmese has been significantly influenced by Pali, the liturgical language of Theravada Buddhism, as well as by Mon and English. Over time, many foreign words entered Burmese as loanwords. After the end of British colonial rule, the government replaced many English terms by creating new Burmese equivalents \cite{multicsd_burmese_2025}. Although Standard Burmese is the official language of Myanmar, widely used by the majority population and more resource-rich than other ethnic languages, it is still considered under-resourced for NLP. It lacks sufficient high-quality data, tools, and benchmarks for effective language processing. However, due to its official status, widespread use, and relative resource availability, we focus on Standard Burmese for benchmarking in this work.

The design of \emph{BURMESE-SAN} focuses on linguistic authenticity and representativeness in vocabulary, sentence structure, and grammar. This enables a comprehensive evaluation of Natural Language Processing (NLP) capabilities in Burmese, encompassing Natural Language Understanding (NLU), Natural Language Reasoning (NLR), and Natural Language Generation (NLG). 
Regarding sentence structure, \emph{BURMESE-SAN} is designed to reflect natural and native use of Standard Burmese. We carefully prepared the evaluation data to help evaluate how well the LLMs handle real-world Burmese language use. (See Section \ref{sec:task_and_dataset}). The benchmark uses commonly spoken vocabulary, focusing on words and phrases frequently used in daily conversation. It also naturally includes loanwords - mainly from Pali and English - that are now a regular part of modern Burmese lexicon.

\subsection{Dataset and Benchmark Overview}

Previous efforts to evaluate Burmese language models have been mostly scattered and uncoordinated. Researchers have developed datasets for Burmese NLP tasks, including text classification, sequence tagging, and machine translation \citep{mysentence, san2024thai-myanmar-nmt, zaw2022english-burmese-pivot, hlaing2022pos-nmt, myocr, myhatespeech}. However, there is still no unified, comprehensive benchmark to systematically evaluate LLMs across different linguistic aspects of the Burmese language.

Recent Southeast Asian benchmarks largely exclude Burmese or rely on machine-translated data. SeaExam and SeaBench \cite{liu-etal-2025-seaexam} introduce localized exam-style and open-ended queries in Vietnamese, Thai, and Indonesian, but do not include Burmese. SailCompass \cite{sailcompass} evaluates Southeast Asian language understanding without any Burmese components. SeaEval \cite{wang-etal-2024-seaeval} provides a multilingual and multicultural benchmark comprising 29 datasets; however, Burmese is excluded, and most of the data are machine-translated or derived from general-purpose multilingual prompts. BHASA covers Indonesian, Tamil, Thai, and Vietnamese \cite{leong2023bhasa}, whereas IndoNLU focuses solely on Indonesian NLU tasks \cite{wilie-etal-2020-indonlu}, lacking evaluations of generative or reasoning capabilities.

Unlike these existing benchmarks, \emph{BURMESE-SAN} provides a \textbf{comprehensive, multi-task evaluation suite for Burmese NLP}, covering NLU, NLR, and NLG tasks. All datasets were carefully curated and adapted to ensure native speaker verification, providing linguistic authenticity, high-quality labels, and cultural relevance. This makes \emph{BURMESE-SAN} the \textbf{first holistic benchmark specifically designed for Burmese}.



\renewcommand{\arraystretch}{1.4}
\begin{table*}[h!]
\centering
\scriptsize
\begin{threeparttable}
\resizebox{\textwidth}{!}{ 
\begin{tabular}{lll llll c}
\toprule
\textbf{Comp.} & \textbf{Task} & \textbf{Dataset} & 
\textbf{Target}\tnote{1} & \textbf{Language} & \textbf{Source}\tnote{2} & 
\textbf{Adaptation} & \textbf{Our contribution} \\
\hline
\multirow{1}{*}{NLU} 
& QA & \href{https://huggingface.co/datasets/facebook/belebele}{Belebele} \cite{bandarkar-etal-2024-belebele} & S (4) & Burmese & English-adapted & Native re-translation & \checkmark \\
& SA & \href{https://huggingface.co/datasets/kornwtp/gklmip-sentiment-mya-classification}{GKLMIP-mya} \cite{sentiment} & L (3) & Burmese & Natively-sourced & Normalization\tnote{3} & \checkmark \\
& TD & \href{https://github.com/ye-kyaw-thu/myHateSpeech}{myHateSpeech} \cite{myhatespeech} & L (9) & Burmese & Natively-sourced & No Adaptation &  \\
\hline
\multirow{1}{*}{NLR} 
& CR & \href{https://huggingface.co/datasets/pkavumba/balanced-copa}{Balanced COPA} \cite{kavumba-etal-2019-choosing} & L (2) & English & English-sourced & Native translation & \checkmark \\
& NLI & \href{https://huggingface.co/datasets/akhtet/myanmar-xnli}{myXNLI} \cite{htet2025myanmar} & L (3) & Burmese & Natively-translated & No Adaptation &  \\
\hline
\multirow{1}{*}{NLG} 
& AS & \href{https://huggingface.co/datasets/csebuetnlp/xlsum}{XL-Sum} \cite{hasan-etal-2021-xl} & Sum & Burmese & English-adapted & Native re-translation & \checkmark \\
& MT & \href{https://huggingface.co/datasets/openlanguagedata/flores_plus}{FLORES+} \cite{nllb-24} & Trans & Burmese & English-adapted & Native re-translation & \checkmark \\
\bottomrule
\end{tabular}
} 
\begin{tablenotes}
    \item[1] Number of options shown in parentheses.  
    \textbf{L}: Label; \textbf{S}: Span; \textbf{Sum}: Summary; \textbf{Trans}: Translate
    \item[2] 
    \textbf{English-adapted}: previously translated from English, then further adapted or corrected by native speakers.  \\
    \textbf{English-sourced}: originally written in English. \\
    \textbf{Natively-sourced}: originally created in Burmese by native speakers.  
    \textbf{Natively-translated}: translated by native speakers
    \item[3] 
    \textbf{Normalization}: spelling was standardized to follow consistent orthography and modern usage norms, \\including Unicode normalization and correction of common typos.
\end{tablenotes}
\vspace{-3mm}
\caption{Dataset information for each task in \emph{BURMESE-SAN}. A checkmark in the \textit{Our contribution} column indicates direct contributions to the adaptation process.}
\vspace{-3mm}
\label{tab:dataset_info_burmese_san}
\end{threeparttable}
\end{table*}


\section{Task and Dataset Curation}
\label{sec:task_and_dataset}

\renewcommand{\arraystretch}{1}
\begin{table}[!ht]
    \centering
    \small
    \begin{tabular}{lllr}
        \toprule
        \textbf{Competency} & \textbf{Task} & \textbf{Label} & \textbf{\# samples} \\
        \hline
        \multirow{1}{*}{NLU} 
        & QA  & Span & 120 \\
        & SA  & Positive & 200 \\
        & & Negative & 200 \\
        & & Neutral & 200 \\
        & TD  & Clean & 200 \\
        & & Hate & 200 \\
        \hline
        \multirow{1}{*}{NLR} 
        & CR  & Cause & 200 \\
        & & Effect & 200 \\
        & NLI  & Contradiction & 200 \\
        & & Entailment & 200 \\
        & & Neutral & 200 \\
        \hline
        \multirow{2}{*}{NLG} 
        & AS  & Summary & 100 \\
        & MT & MYA $\rightarrow$ ENG & 850 \\
        & & ENG $\rightarrow$ MYA & 850 \\
        \hline
        Total & & & 3,920 \\
        \bottomrule
    \end{tabular}
    \vspace{-3mm}
    \caption{Class distribution per task in \emph{BURMESE-SAN} Benchmark Dataset}
    \vspace{-3mm}
    \label{tab:dataset_distribution}
\end{table}
\renewcommand{\arraystretch}{1.0}


As shown in Table~\ref{tab:dataset_distribution}, \emph{BURMESE-SAN} covers tasks across Natural Language Understanding (NLU), Natural Language Reasoning (NLR), and Natural Language Generation (NLG) with balanced class distributions, totaling 3,920 samples. 
We discuss how to formulate \emph{BURMESE-SAN}, starting from task and dataset selection in Sections~\ref{subsec:task_selection} and \ref{subsec:dataset_selection}, to ensure that our benchmark is holistic.
Finally, we then describe the dataset adaptation and annotation process in Section~\ref{subsec:dataset_adapt}, including how we ensure the quality of our benchmark.
Note that more information about the tasks and datasets, along with task descriptions, dataset information, and quality, is discussed further in Appendix \ref{sec:appendix_task_eval}, \ref{sec:appendix_dataset_quality}, and \ref{appendix:challenges}.

\subsection{Task Selection} \label{subsec:task_selection}

Similar to well-exhibited benchmarks in the SEA languages~\cite{montalan2025batayanfilipinonlpbenchmark,liu-etal-2025-seaexam}, we require tasks that are widely used to evaluate the robustness of LLMs that reflect real-world scenarios.
The tasks should consist of classical NLP tasks, such as classification, and LLM-widely adopted tasks, including question answering (QA), summarization (AS), and machine translation (MT), as well as reasoning and understanding tasks, including natural language inference (NLI).
Therefore, \emph{BURMESE-SAN} brings comprehensive LLM evaluation to Burmese with seven tasks across NLU (sentiment analysis, question answering, and toxicity detection), NLG (abstractive summarization and machine translation), and NLR (causal reasoning and natural language inference), as shown in Table~\ref{tab:dataset_info_burmese_san}. This ensures that our benchmark covers all aspects of LLM evaluation.

\subsection{Dataset Selection} \label{subsec:dataset_selection}

To build \emph{BURMESE-SAN}, we collected open-source datasets with clear sources, focusing on those that reflect authentic Burmese language use in domains such as social media, news, and forums.  
For tasks like TD and NLI, high-quality datasets are already available that do not require adaptation. 
These datasets are either formulated in Burmese (monolingual datasets) or carefully edited by previous works to fix translation issues~\cite{myhatespeech, htet2025myanmar}. 

However, for other tasks, existing datasets required translation or editing to meet our goals of creating high-quality and culturally relevant datasets. 
Therefore, we adapted English and partially translated data sets, including FLORES+~\cite{nllb-24}, Belebele~\cite{bandarkar-etal-2024-belebele}, GKLMIP-mya~\cite{sentiment}, XL-Sum~\cite{hasan-etal-2021-xl}, and Balanced COPA~\cite{kavumba-etal-2019-choosing}, to create Burmese-SAN. While these datasets are well-known and widely used in LLM evaluation benchmarks~\cite{lovenia-etal-2024-seacrowd, susanto2025seahelmsoutheastasianholistic}, they were not readily available in Burmese or were inadequately translated. 
We applied normalization and native re-translation to ensure suitability for inclusion in Burmese-SAN.  
Table~\ref{tab:dataset_info_burmese_san} also summarizes the usage, adaptations, and our contributions, highlighting the \emph{comprehensive, native-speaker-verified nature of Burmese-SAN}, which distinguishes it from prior benchmarks \cite{vayani2025languagesmatterevaluatinglmms, lovenia-etal-2024-seacrowd}. 

\subsection{Dataset Adaptation and Annotation}
\label{subsec:dataset_adapt}

As we discussed in the previous discussion, although some Burmese datasets are available, their data format and quality need refinement to make them \emph{natural in Burmese}, \emph{culturally suitable}, and \emph{reliable for evaluation}.
As shown in Figure~\ref{fig:task_summary} (Right), we designed a four-step process (random sampling and filtering, translation, text normalization, and label verification, as well as native revision) to combine Burmese and non-Burmese datasets into a single benchmark.
Note that, to ensure dataset quality, we employed native Burmese speakers (ages 18--25), primarily university students enrolled in international programs as annotators.

\noindent
\textbf{Random Sampling and Filtering}
For each task, we first sampled data and restricted text length to 20–3,000 characters, ensuring balanced class distributions to reduce bias. 
All NLU and NLR tasks were limited to 32 tokens per output, whereas generation tasks utilized longer outputs, with 256 tokens for MT and 512 tokens for AS, to accommodate richer content generation.
Low-quality items, such as duplicates, unclear text, or unnatural writing, were removed by native speakers to maintain the high quality of Burmese examples. 
For the natively sourced datasets (SA, TD) and the natively translated NLI \cite{htet2025myanmar}, we applied task-specific filtering.
%
%
For the SA and TD tasks, we removed sentences with contextually ambiguous expressions that could obscure labels. For the NLI task, we ensured premise–hypothesis pairs accurately reflected the intended relation.

\noindent
\textbf{Translation and Binary Criteria Rating}
For tasks originally in English (QA, CR, AS, MT), the data was translated into Burmese using annotators, ensuring high-quality data. 
For the English-sourced dataset (CR), we translated it from English to Burmese by leveraging our native speaker annotators. 
In contrast, the English-adapted datasets (QA, AS, and MT) were re-translated due to the low quality of existing translations.
%
After translation, our native-speaker annotators applied a binary rating (agree/disagree) to evaluate dataset quality based on multiple criteria. 
We used Completeness, Fluency, and Sensibility to assess information retention, grammaticality, and contextual logic for all tasks; Faithfulness for translation to evaluate meaning preservation; and Relevance and Coherence for summarization to measure content focus and structural flow. The initial overall mean joint agreement across all tasks and criteria is 99.11\%, indicating a consistently high level of annotator alignment throughout the evaluation. 
%
%
For each sample, if it did not receive full agreement from our annotators, it was then revised by them to improve quality. 
This process efficiently filtered out poor-quality translations without requiring full re-annotation, ensuring that only accurate and fluent Burmese text was included in the benchmark. 
%

\noindent
\textbf{Text Normalization and Label Verification}
In the next step, normalization was applied to improve consistency for datasets that were natively sourced (SA and TD). 
This process primarily served to resolve ambiguities and measure inter-annotator agreement, preserving dataset integrity while improving reliability for downstream evaluation.
For Toxicity Detection (TD), despite spelling errors, distorted spellings were preserved, as they reflect real social media usage and intentional attempts to bypass detection. 
For Sentiment Analysis (SA), typos, spelling mistakes, and character-order issues were corrected so that errors would not affect classification. 
Moreover, code-switching (English-Burmese) samples were preserved in both tasks, as it is a natural part of Myanmar social media. 
For the gold label of each sample, due to the subjective nature of these tasks, native speakers verified and, when necessary, adjusted the labels. 
We retained the original labels as far as possible, as our goal was to ensure label consistency and quality rather than redefine the task. 
We found that the agreement between the original and verified labels was very high. For Sentiment Analysis (SA), Cohen’s Kappa = 0.948, Krippendorff’s Alpha = 0.948, and total agreement = 96.5\%. For Toxicity Detection (TD), Cohen’s Kappa = 0.985, Krippendorff’s Alpha = 0.985, and total agreement = 99.25\%.

\noindent
\textbf{Native Revision}
Finally, native speakers conducted a final review of all text. 
This step focused on refining linguistic naturalness, cultural appropriateness, and readability, including minor adjustments to grammar, punctuation, and style. 
The goal was not to alter the meaning, but to ensure that the dataset reflects authentic Burmese language use suitable for downstream tasks.

\begin{table*}[t]
\centering
\resizebox{\textwidth}{!}{%
\begin{tabular}{l r r r r r r r r}
\toprule
\textbf{Model} & \textbf{MY} & \textbf{CR} & \textbf{NLI} & \textbf{QA} & \textbf{SA} & \textbf{TD} & \textbf{AS} & \textbf{MT} \\
\midrule
\multicolumn{9}{l}{\textit{Small Models ($<$ 14B)}} \\
\midrule
MERaLiON 2 (10B) & 10.66 ± 0.16 & 0.00 ± 0.00 & 12.61 ± 0.31 & 0.00 ± 0.00 & 0.07 ± 0.13 & 14.79 ± 0.09 & 19.49 ± 0.27 & 35.33 ± 0.15 \\
Olmo 2 1124 (13B) & 1.84 ± 0.09 & 0.00 ± 0.00 & 0.00 ± 0.00 & 0.00 ± 0.00 & 0.00 ± 0.00 & 0.79 ± 0.03 & 0.00 ± 0.00 & 1.04 ± 0.03 \\
Olmo 2 1124 (7B) & 2.18 ± 0.12 & 0.00 ± 0.00 & 0.00 ± 0.00 & 0.00 ± 0.00 & 0.00 ± 0.00 & 1.80 ± 0.04 & 0.00 ± 0.00 & 6.44 ± 0.10 \\
Olmo 3 (7B) & 3.41 ± 0.11 & 0.00 ± 0.00 & 0.00 ± 0.00 & 0.00 ± 0.00 & 0.00 ± 0.00 & 3.66 ± 0.05 & 14.00 ± 0.14 & 1.11 ± 0.02 \\
Tulu 3 (8B) & 10.95 ± 0.14 & 0.00 ± 0.00 & 1.35 ± 0.13 & 5.89 ± 0.64 & 0.00 ± 0.00 & 13.70 ± 0.06 & \underline{21.48 ± 0.17} & 37.48 ± 0.10 \\
SEA-LION v3-Gemma-2 (9B) & 15.40 ± 0.22 & 0.00 ± 0.00 & 25.10 ± 0.37 & 4.52 ± 1.35 & 0.00 ± 0.00 & 20.64 ± 0.11 & 14.84 ± 0.29 & 53.10 ± 0.15 \\
SEA-LION v3-Llama (8B) & 17.88 ± 0.26 & 4.45 ± 1.29 & 17.01 ± 0.70 & 37.41 ± 0.81 & 22.19 ± 0.97 & 12.79 ± 0.26 & 5.61 ± 0.27 & 13.92 ± 0.12 \\
SEA-LION v4-Apertus (8B) & 16.68 ± 0.22 & 0.00 ± 0.00 & 0.00 ± 0.00 & 23.44 ± 0.97 & 19.05 ± 0.43 & 15.71 ± 0.08 & 5.47 ± 0.13 & 55.46 ± 0.16 \\
SEA-LION v4-Gemma-3-VL (4B) & 26.24 ± 0.13 & 0.00 ± 0.00 & \underline{34.39 ± 0.30} & 47.78 ± 0.00 & \underline{43.11 ± 0.15} & 27.90 ± 0.07 & 17.93 ± 0.10 & \underline{62.56 ± 0.12} \\
SEA-LION v4-Qwen-3-VL (4B) & 23.31 ± 0.17 & 28.80 ± 0.41 & 21.26 ± 0.21 & 42.81 ± 0.25 & 31.54 ± 0.14 & 20.37 ± 0.10 & 0.63 ± 0.03 & 24.90 ± 0.13 \\
SEA-LION v4-Qwen-3-VL (8B) & \underline{30.67 ± 0.19} & \underline{38.77 ± 0.41} & 23.28 ± 0.25 & \underline{58.22 ± 0.60} & 36.21 ± 0.27 & 31.76 ± 0.11 & 15.49 ± 0.21 & 51.70 ± 0.20 \\
Qwen 2.5 (7B) & 8.04 ± 0.15 & 0.00 ± 0.00 & 5.97 ± 0.82 & 0.00 ± 0.00 & 0.00 ± 0.00 & 8.77 ± 0.15 & 19.35 ± 0.18 & 11.22 ± 0.11 \\
Qwen 3 (4B) & 18.61 ± 0.13 & 0.00 ± 0.00 & 29.18 ± 0.48 & 40.59 ± 0.41 & 35.60 ± 0.13 & 21.34 ± 0.09 & 21.48 ± 0.09 & 40.17 ± 0.12 \\
Qwen 3 (8B) & 26.26 ± 0.20 & 30.70 ± 0.65 & 3.16 ± 0.14 & 40.04 ± 0.73 & 33.83 ± 0.33 & 25.73 ± 0.13 & 20.11 ± 0.15 & 47.90 ± 0.13 \\
Qwen 3 VL (4B) & 20.42 ± 0.16 & 22.77 ± 0.57 & 21.53 ± 0.32 & 44.44 ± 0.38 & 36.70 ± 0.13 & 20.01 ± 0.13 & 0.55 ± 0.00 & 34.34 ± 0.11 \\
Qwen 3 VL (8B) & 30.25 ± 0.22 & 36.85 ± 0.59 & 32.05 ± 0.46 & 53.59 ± 0.70 & 36.41 ± 0.26 & \underline{32.60 ± 0.17} & 16.51 ± 0.23 & 49.72 ± 0.09 \\
Babel (9B) & 8.01 ± 0.18 & 0.00 ± 0.00 & 0.00 ± 0.00 & 0.00 ± 0.00 & 22.39 ± 0.63 & 8.64 ± 0.08 & 18.33 ± 0.33 & 17.51 ± 0.22 \\
SeaLLMs V3 (7B) & 7.13 ± 0.16 & 0.00 ± 0.00 & 0.00 ± 0.00 & 0.00 ± 0.00 & 0.00 ± 0.00 & 8.47 ± 0.08 & 17.68 ± 0.31 & 18.87 ± 0.15 \\
Aya Expanse (8B) & 3.03 ± 0.13 & 0.00 ± 0.00 & 0.00 ± 0.00 & 0.00 ± 0.00 & 0.00 ± 0.00 & 1.48 ± 0.04 & 0.00 ± 0.00 & 2.82 ± 0.06 \\
Command R7B 12-2024 (7B) & 3.19 ± 0.13 & 0.00 ± 0.00 & 0.00 ± 0.00 & 0.00 ± 0.00 & 0.00 ± 0.00 & 4.36 ± 0.07 & 10.06 ± 0.28 & 9.73 ± 0.14 \\
Gemma 2 (9B) & 9.63 ± 0.17 & 0.00 ± 0.00 & 9.16 ± 0.63 & 0.00 ± 0.00 & 0.00 ± 0.00 & 13.02 ± 0.13 & 14.08 ± 0.28 & 35.64 ± 0.13 \\
Gemma 3 VL (12B) & \textbf{42.46 ± 0.15} & \textbf{44.93 ± 0.77} & 26.22 ± 0.16 & \textbf{67.11 ± 0.41} & \textbf{48.96 ± 0.22} & \textbf{40.97 ± 0.15} & 18.77 ± 0.26 & \textbf{70.96 ± 0.09} \\
Gemma 3 VL (4B) & 20.56 ± 0.19 & 0.00 ± 0.00 & 22.12 ± 0.31 & 35.74 ± 0.38 & 39.18 ± 0.16 & 22.25 ± 0.09 & 19.35 ± 0.12 & 50.90 ± 0.14 \\
Llama 3 (8B) & 3.20 ± 0.06 & 0.00 ± 0.00 & 0.00 ± 0.00 & 0.00 ± 0.00 & 0.00 ± 0.00 & 4.89 ± 0.03 & 0.00 ± 0.00 & 22.95 ± 0.15 \\
Llama 3.1 (8B) & 8.45 ± 0.19 & 0.00 ± 0.00 & 0.00 ± 0.00 & 7.48 ± 1.02 & 0.00 ± 0.00 & 10.72 ± 0.09 & \textbf{26.73 ± 0.27} & 20.69 ± 0.14 \\
Ministral 2410 (8B) & 3.90 ± 0.13 & 0.00 ± 0.00 & 0.29 ± 0.23 & 0.00 ± 0.00 & 0.05 ± 0.10 & 6.61 ± 0.07 & 4.22 ± 0.19 & 27.35 ± 0.17 \\
Sailor2 (8B) & 11.65 ± 0.13 & 0.00 ± 0.00 & \textbf{43.74 ± 0.79} & 0.26 ± 0.27 & 0.00 ± 0.00 & 19.30 ± 0.14 & 0.00 ± 0.00 & 48.76 ± 0.14 \\
Apertus (8B) & 9.30 ± 0.21 & 0.00 ± 0.00 & 0.25 ± 0.22 & 3.78 ± 1.27 & 2.92 ± 0.74 & 12.13 ± 0.10 & 13.22 ± 0.27 & 40.68 ± 0.16 \\
\midrule
\multicolumn{9}{l}{\textit{Medium Models (14B--32B)}} \\
\midrule
Olmo 2 0325 (32B) & 4.38 ± 0.15 & 0.00 ± 0.00 & 0.00 ± 0.00 & 0.00 ± 0.00 & 0.00 ± 0.00 & 3.91 ± 0.05 & 0.00 ± 0.00 & 14.73 ± 0.12 \\
SEA-LION v4-Gemma-3-VL (27B) & 47.18 ± 0.15 & 57.15 ± 0.47 & 52.92 ± 0.18 & 69.11 ± 0.41 & \textbf{47.56 ± 0.26} & 48.95 ± 0.11 & 11.43 ± 0.23 & \underline{77.83 ± 0.10} \\
SEA-LION v4-Qwen-3-VL (32B) & \textbf{49.56 ± 0.14} & \textbf{66.38 ± 0.20} & \textbf{57.13 ± 0.20} & \textbf{81.52 ± 0.22} & 40.49 ± 0.13 & \textbf{51.70 ± 0.08} & 23.56 ± 0.08 & 64.00 ± 0.19 \\
Qwen 2.5 (14B) & 13.59 ± 0.19 & 0.00 ± 0.00 & 4.36 ± 0.68 & 33.96 ± 0.68 & 0.00 ± 0.00 & 14.25 ± 0.13 & 21.50 ± 0.11 & 32.46 ± 0.10 \\
Qwen 2.5 (32B) & 26.99 ± 0.17 & 13.68 ± 0.64 & 5.60 ± 0.47 & 43.04 ± 0.71 & 32.36 ± 0.33 & 22.78 ± 0.14 & \underline{24.05 ± 0.12} & 39.36 ± 0.15 \\
Qwen 3 (14B) & 32.46 ± 0.13 & 31.75 ± 0.52 & 17.80 ± 0.30 & 61.26 ± 0.63 & 36.96 ± 0.23 & 31.52 ± 0.12 & 22.76 ± 0.12 & 51.87 ± 0.16 \\
Qwen 3 (32B) & 44.60 ± 0.19 & \underline{60.68 ± 0.45} & 52.07 ± 0.34 & \underline{74.48 ± 0.73} & 41.37 ± 0.31 & 45.33 ± 0.12 & \textbf{25.31 ± 0.18} & 44.71 ± 0.18 \\
Qwen 3 VL (32B) & 40.89 ± 0.19 & 49.48 ± 0.48 & 43.49 ± 0.28 & 74.04 ± 0.27 & \underline{47.17 ± 0.17} & 40.26 ± 0.12 & 10.77 ± 0.23 & 56.03 ± 0.11 \\
Aya Expanse (32B) & 6.44 ± 0.15 & 0.25 ± 0.24 & 0.25 ± 0.22 & 0.00 ± 0.00 & 0.00 ± 0.00 & 5.67 ± 0.08 & 0.00 ± 0.00 & 20.66 ± 0.18 \\
Command R 08-2024 (32B) & 4.87 ± 0.15 & 0.00 ± 0.00 & 1.37 ± 0.69 & 1.07 ± 0.61 & 0.00 ± 0.00 & 6.30 ± 0.13 & 16.97 ± 0.15 & 9.75 ± 0.11 \\
Gemma 2 (27B) & 23.97 ± 0.24 & 25.78 ± 1.05 & 32.10 ± 0.27 & 31.30 ± 1.50 & 21.46 ± 0.87 & 29.59 ± 0.21 & 21.41 ± 0.19 & 50.32 ± 0.12 \\
Gemma 3 VL (27B) & \underline{48.14 ± 0.17} & 57.08 ± 0.51 & \underline{53.82 ± 0.16} & 69.85 ± 0.31 & 47.13 ± 0.24 & \underline{49.96 ± 0.12} & 13.16 ± 0.25 & \textbf{79.44 ± 0.07} \\
phi-4 (14B) & 6.45 ± 0.18 & 0.00 ± 0.00 & 0.00 ± 0.00 & 0.00 ± 0.00 & 17.77 ± 0.69 & 6.65 ± 0.09 & 10.79 ± 0.39 & 16.19 ± 0.12 \\
Mistral Small 3.1 2503 (24B) & 2.22 ± 0.11 & 0.00 ± 0.00 & 0.00 ± 0.00 & 0.00 ± 0.00 & 0.00 ± 0.00 & 2.38 ± 0.05 & 3.48 ± 0.09 & 6.34 ± 0.14 \\
Sailor2 (20B) & 8.55 ± 0.11 & 0.00 ± 0.00 & 0.00 ± 0.00 & 0.00 ± 0.00 & 0.00 ± 0.00 & 11.66 ± 0.04 & 0.00 ± 0.00 & 52.86 ± 0.15 \\
\midrule
\multicolumn{9}{l}{\textit{Large Models ($>$ 32B)}} \\
\midrule
Tulu 3 (70B) & 35.11 ± 0.17 & 56.78 ± 0.51 & 43.75 ± 0.43 & 66.85 ± 0.68 & 34.50 ± 0.40 & 43.89 ± 0.14 & 25.97 ± 0.24 & 66.60 ± 0.09 \\
SEA-LION v3-Llama (70B) & 38.21 ± 0.28 & 54.30 ± 0.91 & 44.63 ± 0.47 & 73.30 ± 0.70 & 0.00 ± 0.00 & 43.98 ± 0.20 & 25.26 ± 0.16 & 63.28 ± 0.16 \\
Qwen 3 A3B (30B MoE) & 25.62 ± 0.12 & 0.00 ± 0.00 & 44.07 ± 0.47 & 69.04 ± 0.44 & 31.12 ± 0.29 & 21.90 ± 0.09 & 1.27 ± 0.12 & 34.84 ± 0.12 \\
Qwen 2.5 (72B) & 27.54 ± 0.20 & 5.58 ± 0.22 & 12.41 ± 0.25 & 48.63 ± 0.45 & 39.85 ± 0.22 & 23.51 ± 0.10 & 21.68 ± 0.27 & 46.32 ± 0.14 \\
Qwen 3 A22B (235B MoE) & \underline{54.29 ± 0.16} & \textbf{71.32 ± 0.23} & \textbf{57.42 ± 0.19} & 76.63 ± 0.07 & \textbf{50.32 ± 0.18} & \textbf{56.51 ± 0.08} & 23.36 ± 0.14 & 78.40 ± 0.06 \\
Qwen 3 Next (80B MoE) & 44.88 ± 0.16 & 59.27 ± 0.60 & 46.77 ± 0.22 & 74.89 ± 0.43 & 41.88 ± 0.20 & 46.01 ± 0.12 & 21.66 ± 0.08 & 58.58 ± 0.15 \\
Babel (83B) & 9.87 ± 0.23 & 0.00 ± 0.00 & 0.40 ± 0.32 & 29.89 ± 2.08 & 8.11 ± 1.02 & 7.72 ± 0.10 & 16.56 ± 0.24 & 9.61 ± 0.15 \\
ERNIE 4.5 (21B MoE) & 17.70 ± 0.15 & 0.00 ± 0.00 & 0.00 ± 0.00 & 0.00 ± 0.00 & 0.00 ± 0.00 & 19.67 ± 0.06 & 10.37 ± 0.19 & 72.24 ± 0.18 \\
ERNIE 4.5 (300B MoE) & \textbf{54.68 ± 0.16} & 68.82 ± 0.22 & 40.39 ± 0.15 & \underline{78.52 ± 0.32} & \underline{49.81 ± 0.16} & \underline{54.27 ± 0.08} & 20.84 ± 0.22 & \textbf{86.25 ± 0.05} \\
Command A 03-2025 (111B) & 16.52 ± 0.23 & 0.00 ± 0.00 & 18.57 ± 0.58 & 30.78 ± 1.22 & 16.68 ± 0.75 & 15.71 ± 0.14 & 5.57 ± 0.35 & 37.10 ± 0.19 \\
Command R+ 08-2024 (104B) & 6.61 ± 0.18 & 0.00 ± 0.00 & 0.00 ± 0.00 & 0.00 ± 0.00 & 1.83 ± 0.68 & 9.19 ± 0.07 & 16.00 ± 0.17 & 25.95 ± 0.17 \\
DeepSeek V3 (671B MoE) & 40.87 ± 0.25 & 59.22 ± 0.80 & 16.31 ± 0.79 & 15.78 ± 1.72 & 39.07 ± 0.33 & 41.73 ± 0.17 & 20.22 ± 0.13 & 72.92 ± 0.13 \\
DeepSeek V3.1 (671B MoE) & 51.30 ± 0.23 & 58.92 ± 0.70 & \underline{51.64 ± 0.37} & 70.41 ± 0.60 & 44.17 ± 0.21 & 53.41 ± 0.16 & 21.47 ± 0.11 & \underline{85.86 ± 0.05} \\
Llama 3 (70B) & 13.09 ± 0.17 & 32.92 ± 0.66 & 35.18 ± 0.43 & 15.33 ± 1.12 & 0.00 ± 0.00 & 24.03 ± 0.14 & 0.00 ± 0.00 & 49.89 ± 0.11 \\
Llama 3.1 (70B) & 19.87 ± 0.24 & 0.00 ± 0.00 & 35.08 ± 0.71 & 37.63 ± 1.30 & 0.00 ± 0.00 & 26.76 ± 0.15 & 27.30 ± 0.26 & 58.42 ± 0.20 \\
Llama 3.3 (70B) & 23.07 ± 0.15 & 0.00 ± 0.00 & 42.51 ± 0.31 & 3.70 ± 0.72 & 0.00 ± 0.00 & 29.50 ± 0.09 & \underline{28.32 ± 0.21} & 60.06 ± 0.15 \\
Llama 4 Maverick (400B MoE) & 51.49 ± 0.20 & 67.87 ± 0.21 & 31.87 ± 0.39 & \textbf{80.00 ± 0.00} & 44.92 ± 0.11 & 52.54 ± 0.10 & \textbf{30.66 ± 0.15} & 81.86 ± 0.05 \\
Llama 4 Scout (109B MoE) & 45.54 ± 0.17 & \underline{70.40 ± 0.32} & 31.74 ± 0.21 & 75.93 ± 0.19 & 13.19 ± 0.43 & 49.98 ± 0.09 & 25.53 ± 0.25 & 81.15 ± 0.04 \\
Mistral Large 2411 (123B) & 26.17 ± 0.29 & 7.00 ± 1.42 & 10.76 ± 0.40 & 41.41 ± 1.39 & 36.23 ± 0.41 & 24.14 ± 0.27 & 19.68 ± 0.28 & 55.08 ± 0.20 \\
Kimi K2 Instruct 0905 (1040B MoE) & 43.94 ± 0.19 & 53.02 ± 1.07 & 13.98 ± 0.36 & 58.81 ± 0.94 & 40.21 ± 0.47 & 41.64 ± 0.20 & 23.02 ± 0.20 & 71.96 ± 0.19 \\
Apertus (70B) & 13.09 ± 0.25 & 0.00 ± 0.00 & 0.00 ± 0.00 & 21.78 ± 1.43 & 0.00 ± 0.00 & 16.60 ± 0.09 & 15.17 ± 0.17 & 58.27 ± 0.17 \\
\bottomrule
\end{tabular}%
}
\caption{Performance of instruct models on BURMESE-SAN tasks. Best model for each task per size group is bold, second best is underlined.}
\vspace{-5mm}
\label{tab:it_models}
\end{table*}

\begin{table*}[t]
\centering
\resizebox{\textwidth}{!}{%
\begin{tabular}{l r r r r r r r r}
\toprule
\textbf{Model} & \textbf{MY} & \textbf{CR} & \textbf{NLI} & \textbf{QA} & \textbf{SA} & \textbf{TD} & \textbf{AS} & \textbf{MT} \\
\midrule
\multicolumn{9}{l}{\textit{Small Models ($<$ 14B)}} \\
\midrule
Olmo 3 Think (7B) & 3.85 ± 0.16 & 0.00 ± 0.00 & 0.00 ± 0.00 & 0.00 ± 0.00 & 0.00 ± 0.00 & 5.88 ± 0.06 & 9.99 ± 0.23 & 17.61 ± 0.12 \\
SEA-LION v3.5 R (Llama) (8B) & 22.08 ± 0.20 & 0.00 ± 0.00 & 30.19 ± 0.64 & 54.00 ± 1.22 & 0.00 ± 0.00 & 22.28 ± 0.15 & 11.02 ± 0.32 & 48.28 ± 0.32 \\
Qwen 3 (Thinking) (4B) & \underline{34.78 ± 0.29} & \underline{40.78 ± 1.30} & \underline{31.98 ± 0.70} & \underline{60.37 ± 0.96} & \underline{31.82 ± 0.53} & \underline{35.77 ± 0.28} & \underline{16.29 ± 0.40} & \underline{56.00 ± 0.11} \\
Qwen 3 (Thinking) (8B) & \textbf{37.12 ± 0.22} & \textbf{55.80 ± 0.95} & \textbf{33.45 ± 0.63} & \textbf{65.19 ± 1.23} & \textbf{40.18 ± 0.37} & \textbf{40.22 ± 0.22} & \textbf{18.28 ± 0.18} & \textbf{59.53 ± 0.12} \\
\midrule
\multicolumn{9}{l}{\textit{Medium Models (14B--32B)}} \\
\midrule
Olmo 3 Think (32B) & 5.51 ± 0.16 & 0.00 ± 0.00 & 0.00 ± 0.00 & 0.00 ± 0.00 & 0.00 ± 0.00 & 5.85 ± 0.07 & 8.80 ± 0.19 & 15.28 ± 0.14 \\
QwQ (32B) & 24.66 ± 0.27 & 0.00 ± 0.00 & 19.04 ± 0.69 & 55.96 ± 1.33 & 30.38 ± 0.49 & 18.58 ± 0.15 & 4.17 ± 0.27 & 38.97 ± 0.19 \\
Qwen 3 (Thinking) (14B) & \underline{43.81 ± 0.24} & \underline{61.45 ± 0.97} & 35.47 ± 0.45 & \underline{73.70 ± 1.26} & \underline{38.71 ± 0.35} & \underline{45.07 ± 0.20} & \underline{20.86 ± 0.12} & \textbf{65.04 ± 0.21} \\
Qwen 3 (Thinking) (32B) & \textbf{52.35 ± 0.26} & \textbf{71.32 ± 0.70} & \textbf{48.61 ± 0.54} & \textbf{77.33 ± 0.87} & \textbf{42.32 ± 0.35} & \textbf{48.49 ± 0.19} & \textbf{22.90 ± 0.16} & 43.43 ± 0.32 \\
Reka Flash 3.1 (21B) & 24.51 ± 0.28 & 3.80 ± 1.34 & \underline{39.16 ± 0.65} & 61.00 ± 1.34 & 5.20 ± 0.75 & 26.51 ± 0.28 & 10.91 ± 0.21 & \underline{56.18 ± 0.19} \\
\midrule
\multicolumn{9}{l}{\textit{Large Models ($>$ 32B)}} \\
\midrule
SEA-LION v3.5 R (Llama) (70B) & 44.02 ± 0.26 & 19.65 ± 1.58 & 41.37 ± 0.56 & 75.07 ± 0.67 & 8.83 ± 0.80 & 41.60 ± 0.29 & \textbf{21.55 ± 0.10} & 78.97 ± 0.13 \\
Qwen 3 (Thinking) (235B MoE) & \textbf{66.39 ± 0.22} & \textbf{77.97 ± 0.44} & \textbf{51.53 ± 0.47} & \textbf{82.44 ± 0.58} & \textbf{44.68 ± 0.26} & \textbf{61.37 ± 0.13} & 20.48 ± 0.09 & 85.48 ± 0.03 \\
Qwen 3 (Thinking) (30B MoE) & 53.54 ± 0.19 & 66.55 ± 0.77 & 48.84 ± 0.59 & 76.26 ± 0.78 & 38.92 ± 0.27 & 53.77 ± 0.16 & \underline{21.41 ± 0.15} & 78.70 ± 0.08 \\
Qwen 3 Next (Thinking) (80B MoE) & 57.97 ± 0.18 & 72.50 ± 0.52 & \underline{49.68 ± 0.54} & \underline{80.04 ± 0.66} & 40.75 ± 0.30 & 56.35 ± 0.15 & 20.08 ± 0.12 & 79.88 ± 0.08 \\
DeepSeek V3.1 Thinking (671B MoE) & \underline{59.46 ± 0.23} & 71.65 ± 0.67 & 48.27 ± 0.58 & 78.26 ± 0.86 & 40.91 ± 0.32 & \underline{57.75 ± 0.15} & 21.21 ± 0.13 & \textbf{86.46 ± 0.04} \\
Deepseek R1 0528 (671B MoE) & 57.31 ± 0.30 & \underline{72.57 ± 0.54} & 47.25 ± 0.70 & 66.07 ± 1.29 & \underline{41.44 ± 0.31} & 56.41 ± 0.17 & 18.28 ± 0.19 & \underline{85.71 ± 0.05} \\
GPT OSS (120B MoE mxfp4) & 48.71 ± 0.21 & 58.95 ± 0.78 & 24.25 ± 0.53 & 70.19 ± 0.89 & 39.11 ± 0.43 & 46.00 ± 0.16 & 17.34 ± 0.11 & 78.03 ± 0.08 \\
GPT OSS (20B MoE mxfp4) & 35.25 ± 0.24 & 37.58 ± 1.00 & 13.97 ± 0.93 & 60.07 ± 1.37 & 36.12 ± 0.44 & 34.42 ± 0.26 & 17.17 ± 0.16 & 67.28 ± 0.09 \\
\bottomrule
\end{tabular}
}
\caption{Performance of Reasoning models on BURMESE-SAN tasks. Best model for each task per size group is bold, second best is underlined.}
\vspace{-4mm}
\label{tab:reason_models}
\end{table*}

\begin{table*}[t]
\centering
\resizebox{\textwidth}{!}{%
\begin{tabular}{l r r r r r r r r}
\toprule
\textbf{Model} & \textbf{MY} & \textbf{CR} & \textbf{NLI} & \textbf{QA} & \textbf{SA} & \textbf{TD} & \textbf{AS} & \textbf{MT} \\
\midrule
Opus 4.1 (2025-08-05) & 45.10 ± 0.15 & 71.57 ± 0.35 & 0.00 ± 0.00 & 0.00 ± 0.00 & 48.35 ± 0.19 & 45.66 ± 0.08 & \underline{24.69 ± 0.09} & 87.52 ± 0.03 \\
Sonnet 4 & 40.45 ± 0.30 & 24.33 ± 0.13 & \textbf{86.01 ± 0.03} & 0.00 ± 0.00 & 47.82 ± 0.25 & 43.22 ± 0.15 & \textbf{68.10 ± 0.67} & 0.00 ± 0.00 \\
Gemini 2 Flash & 59.23 ± 0.16 & 76.58 ± 0.43 & 41.30 ± 0.43 & 79.81 ± 0.33 & 41.48 ± 0.24 & 57.78 ± 0.12 & 22.09 ± 0.09 & 88.23 ± 0.04 \\
Gemini 2.5 Flash & \underline{69.34 ± 0.16} & \underline{83.58 ± 0.40} & 61.07 ± 0.46 & \underline{84.15 ± 0.53} & 39.60 ± 0.21 & \underline{66.16 ± 0.10} & 24.11 ± 0.14 & \underline{89.49 ± 0.02} \\
Gemini 2.5 Pro & \textbf{72.35 ± 0.18} & \textbf{85.75 ± 0.38} & \underline{67.55 ± 0.43} & \textbf{85.48 ± 0.53} & 45.56 ± 0.30 & \textbf{68.74 ± 0.11} & 24.21 ± 0.12 & \textbf{90.22 ± 0.02} \\
GPT 4.1 (2025-04-14) & 55.80 ± 0.18 & 68.80 ± 0.37 & 42.80 ± 0.38 & 77.15 ± 0.31 & \underline{50.24 ± 0.27} & 54.05 ± 0.11 & 21.37 ± 0.24 & 79.73 ± 0.08 \\
GPT 4o (2024-11-20) & 51.61 ± 0.20 & 70.77 ± 0.53 & 39.48 ± 0.43 & 75.67 ± 0.66 & \textbf{50.94 ± 0.35} & 52.34 ± 0.13 & 21.97 ± 0.13 & 78.58 ± 0.11 \\
GPT 5 (2025-08-07) & 66.46 ± 0.19 & 79.87 ± 0.66 & 43.63 ± 0.44 & 81.37 ± 0.47 & 45.70 ± 0.30 & 60.16 ± 0.13 & 17.09 ± 0.07 & 87.46 ± 0.04 \\
\bottomrule
\end{tabular}%
}
\caption{Performance of Commercial models on BURMESE-SAN tasks. Best model for each task per size group is bold, second best is underlined.}
\label{tab:commercial_models}
\vspace{-5mm}
\end{table*}

\section{Experimental Setup}

\subsection{Prompt Templates}
For \emph{BURMESE-SAN}, we designed task-specific prompt templates entirely in Burmese. These templates were carefully aligned with the principles of prompt design established in SEA-HELM \cite{susanto2025seahelmsoutheastasianholistic} to ensure consistency between tasks. In particular, we translate the task prompts from SEA-HELM into the native Burmese language. This is because the design prompt of SEA-HELM maintains a clear separation between instruction and content, as the formal prompts are unambiguous and standardized, thereby reducing confounding factors that might influence model performance due to differences in prompt style rather than genuine understanding of the Burmese text.

\subsection{Evaluation Setup}
\textbf{Tasks.} Each task in BURMESE-SAN, as described in Section \ref{sec:task_and_dataset}, is designed to thoroughly test how well LLMs understand and use the Burmese language. 
A task includes a set of test and example cases, each with an input, reference answers, metadata, and the correct label.
Metrics We adopt multiple evaluation metrics to assess model performance across tasks. For each metric, the model output is treated as the prediction, while the corresponding instance label serves as the reference. 
For NLU and NLR tasks, we report the accuracy score. For machine translation, we use MetricX-24 \cite{juraska-etal-2024-metricx}. For abstractive summarization, we report ROUGE-L F1 using the multilingual ROUGE implementation from XL-Sum \cite{hasan-etal-2021-xl}. 
All metric scores are normalized to a common [0,100] scale following the SEA-HELM normalization process, which accounts for differences in metric ranges, task difficulty, and random baselines (e.g., MetricX originally ranges from [0,25] but is rescaled to [0,100]). 
Each model is evaluated across eight independent runs without greedy decoding (i.e., temperature = 0) (Miller, 2024), and we report the mean normalized score to ensure stable and reliable performance estimates.

\subsection{Models}

\textbf{Model Selection.} We evaluated a set of LLMs across diverse families, parameter scales, and access types. 
The selection includes (i) instruction-tuned models such as Qwen 3 \cite{yang2025qwen3technicalreport, bai2025qwen3vltechnicalreport}, Llama-3/3.1/3.3/4 series \cite{grattafiori2024llama3herdmodels}, Gemma 2/3 \cite{gemmateam2025gemma3technicalreport}, SEA-LION \cite{ng-etal-2025-sea}, DeepSeek-V3, V3.1 (671B MoE) \cite{deepseekai2025deepseekv3technicalreport}, OLMo \cite{groeneveld-etal-2024-olmo}, Mistral \cite{jiang2023mistral7b}, Sailor2-Chat \cite{dou2025sailor2sailingsoutheastasia}, Babel \cite{zhao2025babelopenmultilinguallarge}, and Command-R series; (ii) reasoning-focused models such as DeepSeek-V3.1 Thinking and DeepSeek-R1 \cite{deepseekai2025deepseekr1incentivizingreasoningcapability}, Qwen3-Thinking \cite{yang2025qwen3technicalreport}, QwQ \cite{qwq32b}, GPT-OSS \cite{openai2025gptoss120bgptoss20bmodel}, OLMo-3-Think \cite{olmo2025olmo3}, and SEA-LION v3.5 R \cite{ng-etal-2025-sea}; and (iii) commercial models including Google's Gemini-2/2.5 (Flash and Pro) \cite{geminiteam2025geminifamilyhighlycapable}, OpenAI's GPT-4o \cite{openai2024gpt4ocard}, GPT-4.1 \cite{openai2024gpt4technicalreport}, and GPT-5 \cite{singh2025openaigpt5card}, and Anthropic's Claude family (Sonnet 4 and Opus 4.1) \cite{anthropic2025claude4card}. 
This diverse collection ensures representative coverage of both open-source and proprietary ecosystems and enables analysis of scaling trends from small to large models.

\noindent \textbf{Inference Details.} During evaluation, input prompts are constructed by combining evaluation instances with their corresponding prompt templates. 
The default evaluation setting in BURMESE-SAN for instruction-tuned models is zero-shot prompting, where prompts do not include any in-context input–label examples. 
For decoding parameters, model-specific default configurations are used when available. 
For any unspecified parameters, we apply vLLM default settings. Given the input prompts and decoding parameters, the model generates outputs for evaluation.

\begin{figure*}[h!]
    \centering
    \includegraphics[width=\linewidth]{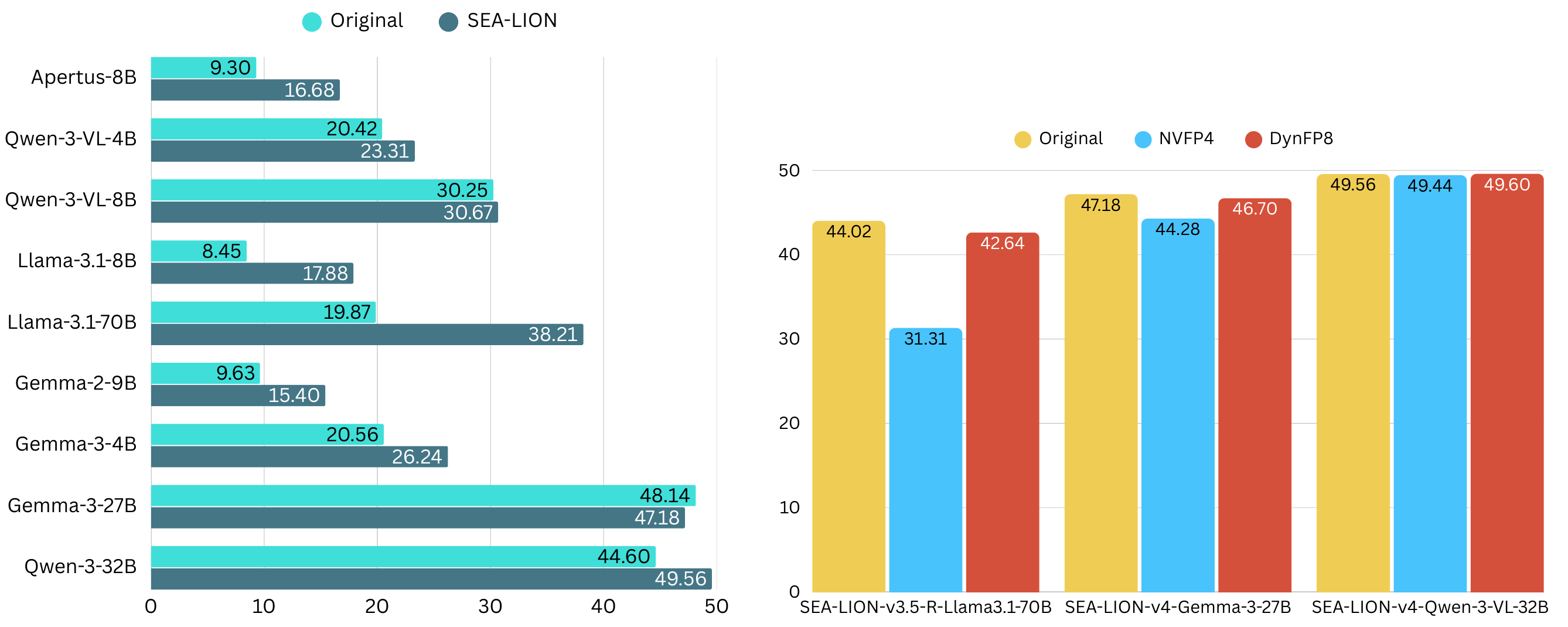}
    \vspace{-5mm}
    \caption{Left: Comparison of original models against SEA-fine-tuned variants, and Right: SEA-LION models with their quantized versions - NVIDIA FP4 (NVFP4) and Dynamic FP8 (DynFP8).}
    \label{fig:comparison}
    \vspace{-5mm}
\end{figure*}

\section{Evaluation Results}
\label{sec:results}

We present our findings organized around five research questions that examine key aspects of Burmese language model capabilities as described in Section \ref{sec:intro}. Tables \ref{tab:it_models}, \ref{tab:reason_models}, and \ref{tab:commercial_models} report performance across all NLP tasks, where the \textit{MY} column denotes overall performance. Figure \ref{fig:comparison} compares original models with their SEA-fine-tuned variants (left) and SEA-LION models with their quantized versions (right), highlighting the effects of regional fine-tuning and quantization.

\noindent \textbf{Finding \#1: Commercial models consistently outperform open-weight models (RQ1).}
Commercial models achieve substantially higher performance on Burmese tasks, led by Gemini 2.5 Pro (72.35\%), Gemini 2.5 Flash (69.34\%), and GPT-5 (66.46\%). In contrast, the strongest open-weight models - ERNIE 4.5 (54.68\%), Qwen 3 235B MoE (54.29\%), and Llama 4 Maverick (51.49\%) - lag behind, with a gap of approximately 17.67\% between the top commercial and open-weight models. This disparity is especially pronounced in tasks requiring cultural and language-specific reasoning.

\noindent \textbf{Finding \#2: Larger models tend to perform better, but scale alone is insufficient (RQ2).}
Performance generally improves with model scale, though gains are non-linear and diminish at larger sizes. While larger variants within families such as Qwen 3 and Gemma 3 outperform smaller ones, notable exceptions exist: DeepSeek V3.1 substantially exceeds V3 (51.30\% vs.\ 40.87\%), and the much larger Kimi K2 Instruct (1040B MoE, 43.94\%) underperforms smaller models. These results highlight the critical role of architecture, training data, and tuning strategies beyond parameter count.

\noindent \textbf{Finding \#3: Southeast Asian fine-tuning benefits certain model families (RQ3).}
SEA fine-tuning yields selective improvements that depend strongly on the base model. While Qwen-based SEA-LION v4 (32B) shows moderate gains (+4.96\%) and the Gemma-based variant slightly degrades (--0.96\%), Llama-based SEA-LION models benefit substantially from SEA fine-tuning, particularly at larger scales. For example, SEA-LION v3 (Llama 3.1) 70B improves markedly over its base counterparts, indicating that SEA-specific data is especially effective for Llama architectures. Task-wise, gains are most pronounced in machine translation (+19.29\%) and question answering (+7.04\%), highlighting the value of regional fine-tuning for cross-lingual and generation-centric tasks.

\noindent \textbf{Finding \#4: Careful quantization preserves performance for most tasks (RQ4).}
Modern quantization methods can largely retain model performance when applied conservatively. For SEA-LION v4 (Qwen) 32B, both 8-bit DynFP8 and 4-bit NVFP4 quantization yield results comparable to full precision. DynFP8 similarly preserves performance for Gemma- and Llama-based models, whereas aggressive NVFP4 quantization causes notable degradation, particularly for reasoning-intensive models. These results indicate that quantization effectiveness depends on both the chosen method and the target task.

\noindent \textbf{Finding \#5: Burmese language capability has improved rapidly across model generations (RQ5).}
Burmese performance has increased substantially across successive model generations, with clear recent acceleration. Major open-weight families such as Llama, Qwen, and Gemma exhibit large generational gains (e.g., Llama 3.3 70B at 23.07\% to Llama 4 Maverick at 51.49\%), while commercial models show steady progress from GPT-4o (51.61\%) to GPT-5 (66.46\%) and from Gemini 2 Flash (59.23\%) to Gemini 2.5 Pro (72.35\%). Similar trends are observed within the SEA-LION series, where newer releases consistently outperform earlier versions; for instance, SEA-LION v4 (Gemma 3) 4B achieves 26.24\%, substantially exceeding SEA-LION v3 (Gemma 2) 9B at 15.40\%.

Commercial models continue to achieve the strongest performance, but the gap with open-weight models is steadily narrowing as architectures and training strategies improve. While model scale remains relevant, performance depends on architectural design, data quality, and instruction tuning rather than parameter count alone. 

Regional fine-tuning yields model-dependent benefits - particularly for Llama-based models. Finally, temporal analysis highlights rapid and consistent improvements in Burmese language capability across model generations, offering practical guidance for model selection and deployment under diverse constraints.

\section{Conclusion}
\label{sec:conclusion}

We introduce \emph{BURMESE-SAN}, the first comprehensive benchmark for evaluating large language models on Burmese across NLU, NLR, and NLG tasks, constructed with high-quality, linguistically natural data spanning diverse domains.

Our evaluation reveals clear performance gaps between model families and generations, demonstrating that Burmese capability is strongly influenced by model architecture, instruction tuning, and training strategy rather than scale alone. In particular, Southeast Asian fine-tuned models - especially SEA-LION variants - consistently improve performance over generation while recent architectural advances such as MoE and reasoning-focused training further accelerate progress. Although commercial models currently achieve the highest overall scores, our results indicate that carefully tuned open-weight models can significantly narrow this gap, especially as Burmese-focused data and training strategies continue to improve. 

Together, these findings underscore the importance of language-specific adaptation and position \emph{BURMESE-SAN} as a robust foundation for future research, evaluation, and deployment of LLMs for Burmese and other low-resource languages.

\section*{Acknowledgement}
This research is supported by the National Research Foundation, Singapore, under its National Large Language Models Funding Initiative. Any opinions, findings, and conclusions or recommendations expressed in this material are those of the author(s) and do not reflect the views of the National Research Foundation, Singapore. The authors also would like to express their sincere gratitude to the annotators and quality control contributors for their careful work and expertise. We also thank our internship students from King Mongkut’s University of Technology Thonburi, Htoo Myat Min Bo and Wira Ye Yint, for their valuable assistance with data annotation and quality assurance.

\section*{Limitations}
\label{sec:limitations}

\paragraph{Benchmarking with formal written Burmese prompt templates only} In this work, although the evaluation data are natural and conversational, we focus on evaluating models using formally written native Burmese prompt templates to ensure clarity, consistency, and grammatical correctness. 
Model performance on informal-style prompt templates may differ from what is observed in our benchmark. Extending \emph{BURMESE-SAN} to include spoken and colloquial prompt templates could be a valuable direction for future work, enabling a more comprehensive evaluation of Burmese language models across different registers and contexts.

\paragraph{Focus on Standard Burmese} We focus our study on the relatively better-resource standard Burmese from central region of Myanmar and do not include other notable dialects such as Arakanese (Rakhine) in the southwest, Tavoyan in the southeast, and Intha in the east, with others such as Yaw, Merguese (Myeik), and Palaw. Future work might explore translation or other community-driven data collection initiatives to extend coverage to dialects of Burmese language.

\section*{Ethical Considerations}
\label{sec:ethics}

Our work on Burmese language technologies contributes to addressing key challenges in linguistic inclusion and improving technological accessibility for underrepresented language communities. The benchmark deliberately incorporates assessments of culturally specific knowledge to mitigate known biases in large language models. This emphasizes the importance of evaluating models within their cultural context rather than assuming universal applicability. Datasets included in \emph{BURMESE-SAN} are from publicly accessible sources and the authors check and include the License information of the datasets in the study.

For benchmark dataset quality assurance, a team of Burmese native speakers was involved in reviewing and annotating the data. The team, composed of university internship students, was recruited through faculty channels. Workload and compensation were communicated in advance and adhered to the university guidelines and regulatory requirements. Given the nature of the toxicity detection (TD) task, the authors and quality assurance team were exposed to potentially offensive material. Measures were taken to mitigate harm: annotators were encouraged to report inappropriate content and had the option to discontinue their work at any time during label verification and native revision.

We do not anticipate any negative social impacts arising from this study. The \emph{BURMESE-SAN} dataset and accompanying codebase will be released under the Creative Commons Attribution Share-Alike 4.0 (CC-BY-SA 4.0) license.

\nocite{*}
\section{Bibliographical References}\label{sec:reference}
\bibliographystyle{lrec2026-natbib}
\bibliography{lrec2026-example}

\nociteLR{*}
\bibliographystyleLR{lrec2026-natbib}
\bibliographyLR{languageresource}

\appendix
\onecolumn

\section{Overview of Tasks and Datasets}
\label{sec:appendix_task_eval}
As described in Section~\ref{sec:task_and_dataset}, we selected seven tasks for inclusion in \emph{BURMESE-SAN}. Table~\ref{tab:dataset_info_burmese_san} lists the source datasets, their adaptations, and usage, all consistent with the original intent.
\begin{itemize}
    \item{\textbf{Abstractive Summarization (AS)}} In this task, an LLM is given a paragraph and must generate a concise sentence summarizing its content. The evaluation focuses not only on identifying the key information but also on paraphrasing the content coherently. We use XL-Sum \cite{hasan-etal-2021-xl} for this task, which contains annotated article-summary pairs.
    \item{\textbf{Causal Reasoning (CR)}} This task requires the LLM to identify the causal relationship between events. Given a premise and a set of statements, the model must determine which statement represents the cause or effect of the premise. We translate from scratch and employ Balanced COPA \cite{kavumba-etal-2019-choosing}, designed to evaluate commonsense causal reasoning with paired alternatives.
    \item{\textbf{Machine Translation (MT)}} Here, an LLM is provided with text in one language and is expected to translate it into another language. In this work, we evaluate both \textit{English to Burmese} and \textit{Burmese to English} translations using the FLORES+ dataset \cite{nllb-24}, which includes translations across multiple languages and domains.
    \item{\textbf{Natural Language Inference (NLI)}} This classification task requires the LLM to determine the relationship between two sentences (X and Y) as one of the following: (a) X implies Y, (b) X contradicts Y, or (c) X neither implies nor contradicts Y. We use myXNLI \cite{htet2025myanmar}, which provides human-annotated examples for cross-lingual inference evaluation.
    \item{\textbf{Question Answering (QA)}} In this task, an LLM is given a passage and a question and must select the span from the passage that answers the question. We use Belebele \cite{bandarkar-etal-2024-belebele}, a multiple-choice reading comprehension dataset designed to evaluate passage understanding.
    \item{\textbf{Toxicity Detection (TD) and Sentiment Analysis (SA)}} Both tasks involve analyzing natural language text. TD requires detecting hate speech or abusive language, while SA involves classifying sentiment as positive, negative, or neutral. We use myHateSpeech \cite{myhatespeech} for TD and GKLMIP-mya \cite{sentiment} for SA.
\end{itemize}

\renewcommand{\arraystretch}{1}
\begin{table}[h!]
\centering
\begin{tabular}{llc}
\toprule
\textbf{Task} & \textbf{Dataset} & \textbf{License} \\
\midrule
QA & Belebele \cite{bandarkar-etal-2024-belebele} & CC BY-NC 4.0 \\
SA & GKLMIP-mya \cite{sentiment} & Unknown \\
TD & myHateSpeech \cite{myhatespeech} & CC BY-NC-SA 4.0 \\
CR & Balanced COPA \cite{kavumba-etal-2019-choosing} & CC BY 4.0 \\
NLI & myXNLI \cite{htet2025myanmar} & CC BY-NC 4.0 \\
AS & XL-Sum \cite{hasan-etal-2021-xl} & CC BY-NC-SA 4.0 \\
MT & FLORES+ \cite{nllb-24} & CC BY-SA 4.0 \\
\bottomrule
\end{tabular}
\caption{Licenses for all datasets included in \emph{BURMESE-SAN}, with corresponding references.}
\vspace{-4mm}
\label{tab:dataset_licenses_burmese_san}
\end{table}

\newpage
\onecolumn
\section{Dataset Quality Assurance}
\label{sec:appendix_dataset_quality}

To ensure dataset quality, we employed bilingual native speakers (ages 18--25), primarily university students enrolled in international programs in Thailand.


\begin{figure}[H]
    \centering
    \includegraphics[width=0.9\linewidth]{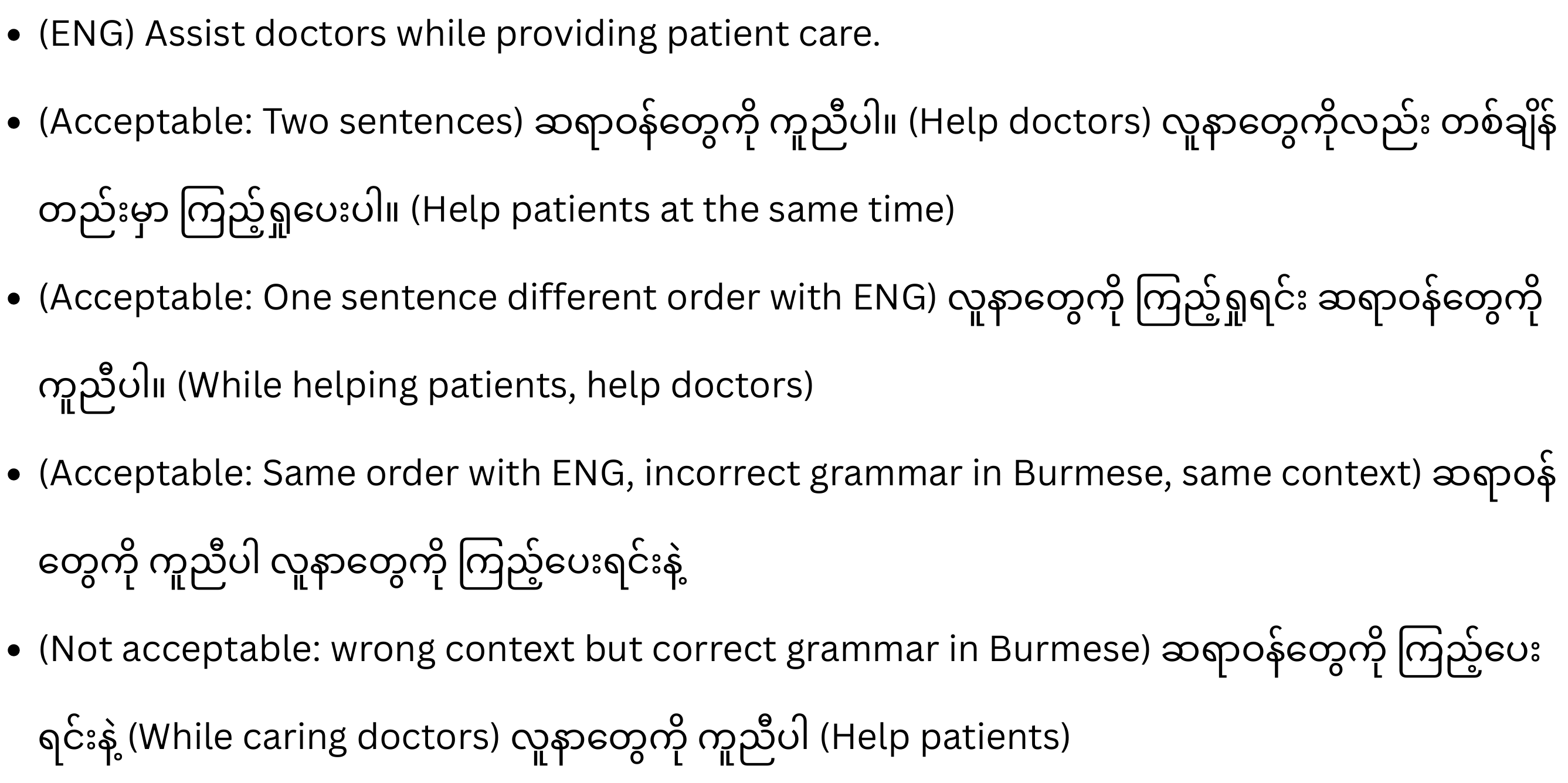}
    \vspace{-4mm}
    \caption{Acceptable and Not Acceptable Grammar Errors in the Dataset.}
    \vspace{-2mm}
    \label{fig:grammer_error}
\end{figure}

\begin{figure}[H]
    \centering
    \includegraphics[width=\linewidth]{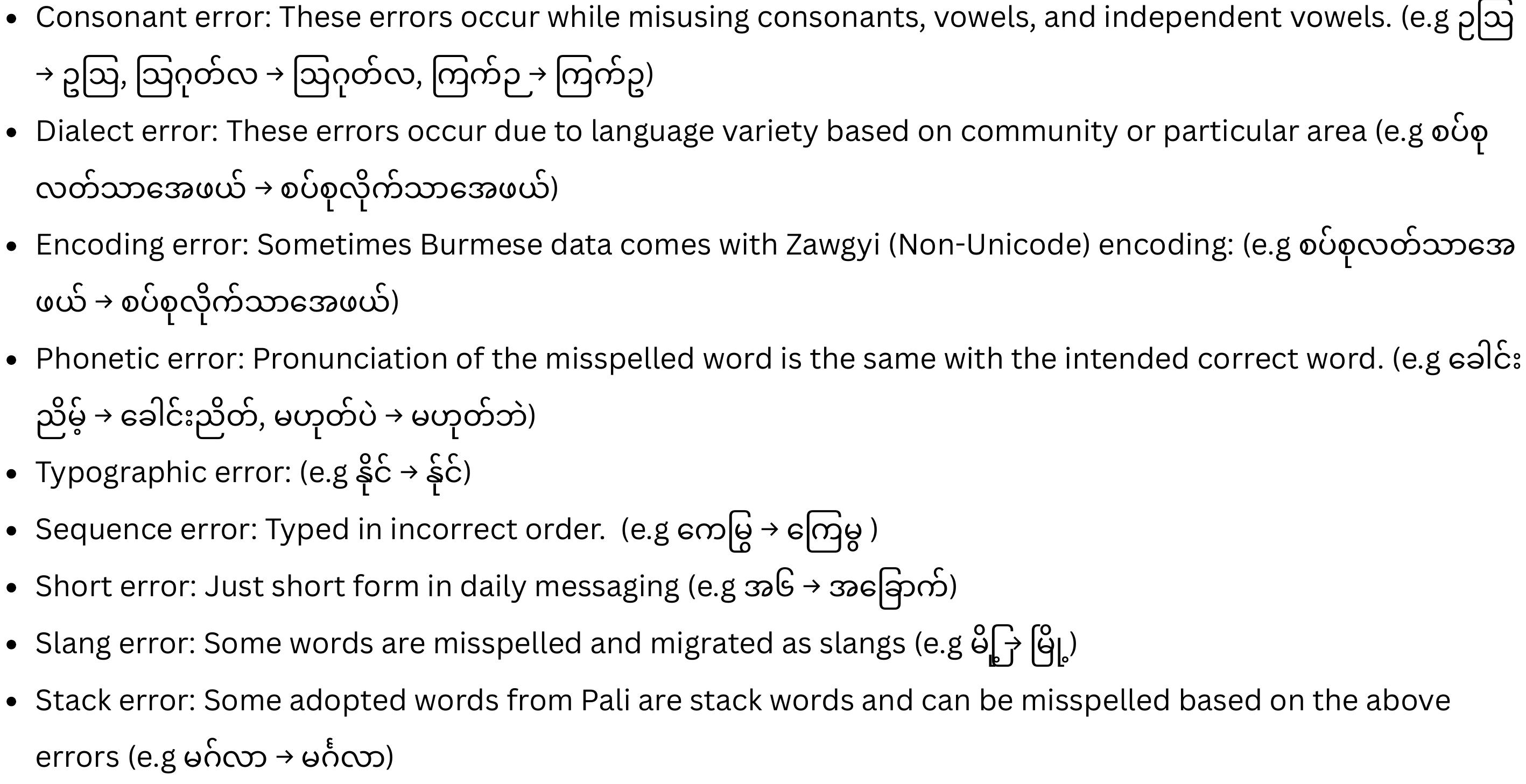}
    \vspace{-4mm}
    \caption{Different Types of Spelling Errors.}
    \vspace{-2mm}
    \label{fig:spelling_error}
\end{figure}

Each task may include different types of linguistic issues. QC members need to fix Grammar and Spelling errors, and the fixed datasets are later used for evaluating LLMs. Figure \ref{fig:grammer_error} and \ref{fig:spelling_error} shows acceptable and unacceptable grammar errors and spelling errors examples. For the grammatical references, we used official guideline published by \citealt{myanmar2013burmesegrammar} and spoken grammar written by \citealt{myanmarspokengrammar}. 

\begin{table}[h!]
\centering
\small
\begin{tabular}{p{3cm} p{3cm} p{9cm}}
\toprule
\textbf{Criteria} & \textbf{Tasks} & \textbf{Definition} \\
\midrule
Completeness & All & Is all the intended information from the source instruction retained in the translated instruction? \\
Fluency & All & Is the output grammatically correct and natural in Burmese? \\
Sensibility & All & Is the translation logical and sensible based on the context of the original instruction? \\
Faithfulness & Translation & Does the translated instruction stay true to the meaning of the original instruction? \\
Relevance & Summarization & Does the output (summary) contain only the most important content? \\
Coherence & Summarization & Is the output (summary) logically structured and easy to follow given the context of the original instruction? \\
\bottomrule
\end{tabular}
\caption{Quality Evaluation Criteria for \emph{BURMESE-SAN}}
\vspace{-2mm}
\label{tab:quality_criteria}
\end{table}

To ensure the quality and reliability of the translated datasets, \textbf{three} QC members evaluated the English–Burmese text pairs for the QA, MT, AS, and CR tasks. Each member assigned a score of 0 (Disagree) or 1 (Agree) for each evaluation criterion, which varied by task. The definition of criteria are as shwon in the Table ~\ref{tab:quality_criteria}. 

\renewcommand{\arraystretch}{1}
\begin{table}[h!t]
    \centering
    \small
    \begin{tabular}{ll>{\scriptsize}r}
        \toprule
        \textbf{Task} & \textbf{Criterion} & \textbf{Joint Agreement (\%)} \\
        \midrule
        \multirow{3}{*}{QA} 
        & Completeness & 100.00 \\
        & Fluency & 97.50 \\
        & Sensibility & 93.33 \\
        \midrule
        \multirow{5}{*}{MT} 
        & Completeness & 100.00 \\
        & Fluency & 99.76 \\
        & Sensibility & 99.88 \\
        & Grammaticality & 99.76 \\
        & Faithfulness & 99.88 \\
        \midrule
        \multirow{6}{*}{AS} 
        & Completeness & 100.00 \\
        & Fluency & 100.00 \\
        & Sensibility & 100.00 \\
        & Relevance of Summary & 100.00 \\
        & Fluency of Summary & 100.00 \\
        & Coherence of Summary & 100.00 \\
        \midrule
        \multirow{3}{*}{CR} 
        & Completeness & 100.00 \\
        & Fluency & 97.00 \\
        & Sensibility & 97.75 \\
        \bottomrule
    \end{tabular}
    \caption{Joint agreement (\%) for each evaluation criterion across tasks requiring (re-)translation. Scores are before revision by native speakers.}
    \vspace{-2mm}
    \label{tab:joint_agreement}
\end{table}

In addition to dataset quality checks, QC members responsible for translation were instructed to focus on the following issues:

\begin{compactitem}
    \item \textbf{Literal Translation:} Avoid overly direct word-for-word rendering that neglects the target language’s natural usage and style.  
    \item \textbf{Cultural Mismatch:} Identify translations that sound unnatural due to irrelevant or inappropriate cultural references.  
    \item \textbf{Incomplete Translation:} Ensure that no parts of the source text are omitted or skipped.  
    \item \textbf{Misinterpretation:} Verify that the intended meaning of the source text is accurately preserved in the translation.  
\end{compactitem}

After evaluation, we revised all translated datasets by reviewing samples that did not receive full agreement among the three annotators. As shown in Table~\ref{tab:joint_agreement}, most criteria achieved high agreement scores - many reaching 100\% - while a few, particularly for Fluency and Sensibility in QA and CR tasks, were slightly lower. These scores reflect the results after re-translation but before manual revision. To ensure high quality and consistency, samples without full agreement were further revised by native speakers. The final dataset contains only samples with full consensus across all evaluation criteria. Dataset statistics, including class distribution are provided in Table~\ref{tab:dataset_distribution}.

\newpage
\onecolumn
\section{Challenges with Developing a Burmese Benchmark} \label{appendix:challenges}
\vspace{-3mm}
\subsection{Issues with Native-Sourced and Native-Translated Data}

\begin{figure}[h!]
    \centering
    \includegraphics[width=0.8\linewidth]{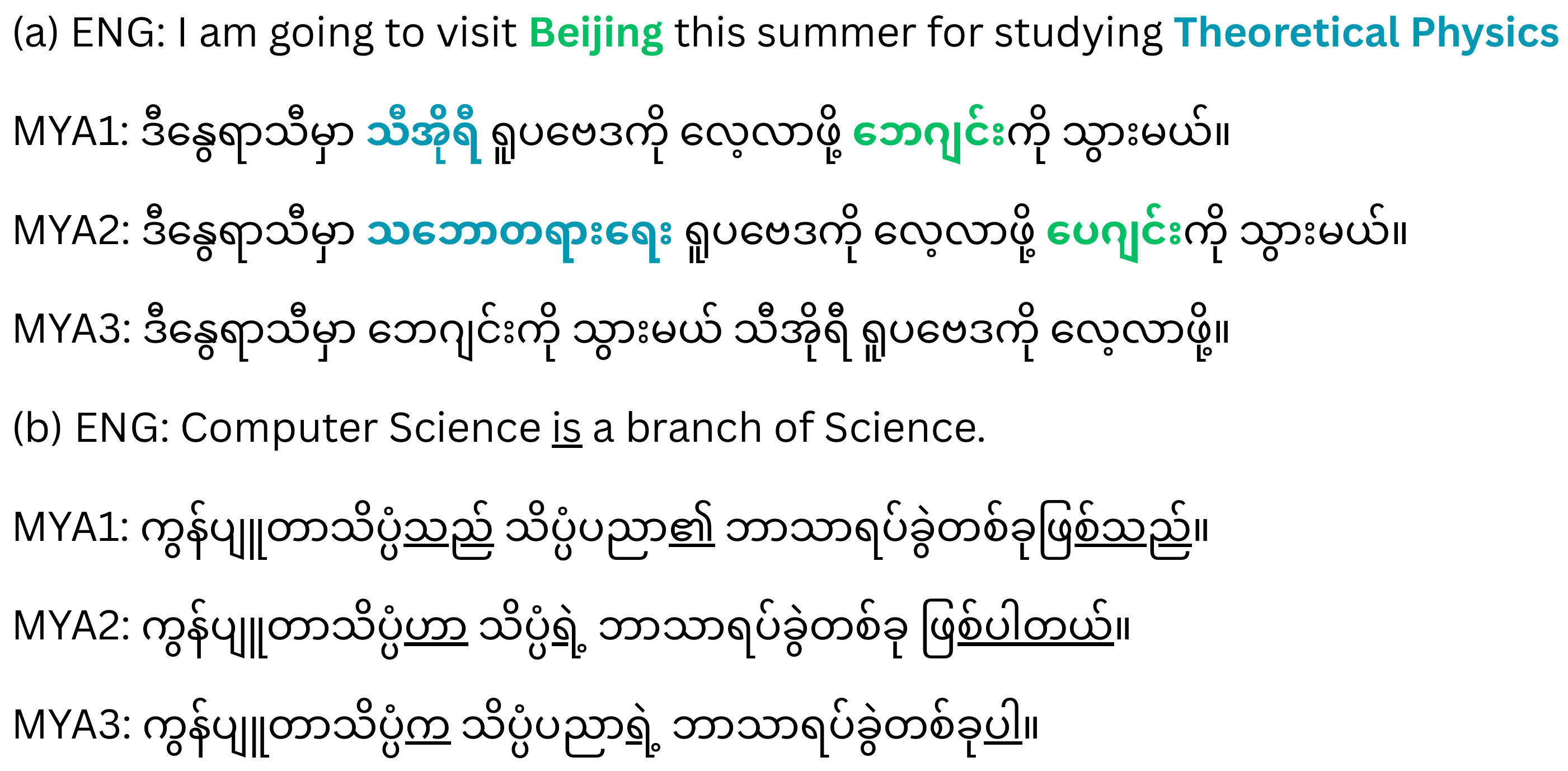}
    \caption{Examples of variation in Burmese translations by native speakers. (a) Differences in technical term usage, transliteration, and word order. (b) Differences in particle choice and formality.}
    \vspace{-4mm}
    \label{fig:burmese_lang_example}
\end{figure}

\paragraph{Inconsistency in Technical Terms}
A common challenge we encountered in Burmese datasets, especially those created or translated by native speakers, is the inconsistent use of loanwords and technical terms. While a Burmese dictionary for scientific and technical vocabulary exists \cite{myanmar2019sci}, it is not widely adopted, and many modern terms are missing. As shown in Figure~\ref{fig:burmese_lang_example} (example a), different translators use different styles. For example, MYA1 and MYA2 provide different translations for the word ``Theoretical'' and use different transliterations for the city name ``Beijing.''

\paragraph{Inconsistency in Syntax}
We encounter another problem, which is syntax inconsistency. 
As shown in \ref{fig:burmese_lang_example}(a), MYA3’s translation follows a similar structure to MYA1 but uses a different word order. Although the syntax is not strictly correct, the meaning remains understandable. MYA3 places the phrase ``I am going to visit Beijing'' before ``for studying Theoretical Physics'' to emphasize the trip itself. In Figure~\ref{fig:burmese_lang_example} (example b), the translations also vary in particle usage. Burmese has multiple particles with similar meanings, and different translators make different stylistic choices. For instance, MYA2 and MYA3 use colloquial language, whereas MYA1 employs formal expressions.

These examples illustrate the broader challenges of working with native-sourced and native-translated Burmese data. The lack of commonly followed standards leads to inconsistencies in terminology. For the syntax, although there are writing standards for both formal and informal styles, the standards are rarely followed. Burmese is a rich and flexible language, and in practice, many speakers do not strictly adhere to formal grammar or standardized syntax, especially in informal contexts like social media. This makes it difficult to ensure consistency, even among native speakers. Such variation can impact the quality and reliability of datasets, particularly for downstream tasks like machine translation or causal reasoning. 
To address this, we conducted a cross-revision process by native speakers reviewing and refining each others' annotations, or translations to improve consistency, clarity, and alignment with the original meaning from original native-sourced or original native-translated datasets. With this approach, we ensure the high-quality and consistency of our datasets, which reflect the real-world application.

\subsection{Issues with English-sourced and English-adapted Data}

English-sourced and English-adapted datasets were initially translated using either semi-automatic methods or manual translation. However, we found that automatic translations, such as those from Google Translate, while grammatically correct, were often unnatural, disfluent, and showed signs of translation.
Many previous works also demonstrate the failure in low-resource machine translations, which results in low-quality~\cite{frontull2025compensatingdatareasoninglowresource, court-elsner-2024-shortcomings} or a lack of localization~\cite{song2025smalllanguagemodelsilver}.
To address this, we applied retranslation to ensure the samples sounded more natural and native. 

For the English-sourced dataset, Balanced COPA~\cite{kavumba-etal-2019-choosing}, there is no Burmese translation in the previous works. Therefore, we need to translate the dataset to Burmese. We used Google Translate, but several issues arose due to literal translation of idioms, incorrect word choices, and reversed meanings. These issues impacted the semantic fidelity of the dataset, especially in the Balanced COPA translation sourced from English. Therefore, we translated the corpus with a bilingual native speaker manually and rated the translation quality, then revised the translation errors again following the procedures discussed in Section \ref{subsec:dataset_adapt} and Appendix \ref{sec:appendix_dataset_quality}. These problems highlight the need for culturally and contextually aware translations. 

Similar issues were found in the original translated version of English-adapted datasets like Flores+ and Belebele, where mistranslations, with unclear question intents and word-for-word translations, often lead to loss of meaning, especially in culturally sensitive or nuanced cases. In another English-adapted dataset, XL-Sum, in the original translation, we identified several quality issues, including inaccurate summaries, missing key information, inconsistent or misleading titles, and incomplete articles. These problems may affect the factual reliability and coherence of the data, highlighting the need for thorough human validation.

Despite re-translation efforts, some samples still did not fully meet fluency and sensibility criteria, as shown in Table~\ref{tab:joint_agreement}. To ensure high-quality and consistent samples for evaluation, we conducted additional manual revisions. In \emph{BURMESE-SAN}, we addressed these challenges through a rigorous translation and revision process in collaboration with native speakers.

\newpage
\onecolumn
\section{Dataset Examples Prompt Templates, and Models Evaluated}
\label{sec:dataset_examples}

\begin{figure}[H]
    \centering
    \includegraphics[width=\linewidth]{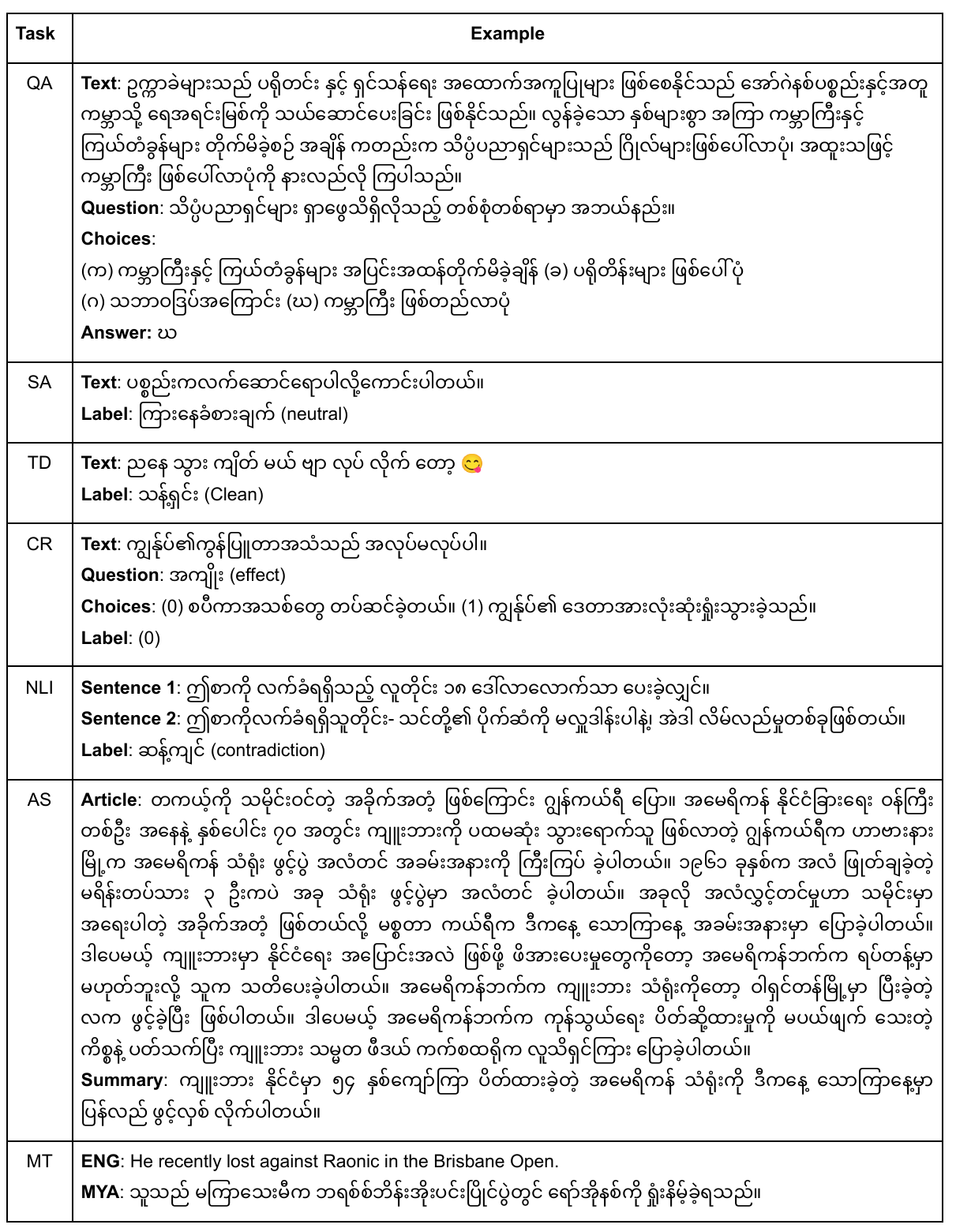}
    \caption{Example Data Samples for each task in \emph{BURMESE-SAN}.}
    \vspace{-4mm}
    \label{fig:dataset_examples}
\end{figure}

\begin{figure}[ht!]
\centering
\begin{minipage}{\textwidth}
\centering

\begin{subfigure}[t]{0.48\textwidth}
    \centering
    \includegraphics[width=\linewidth]{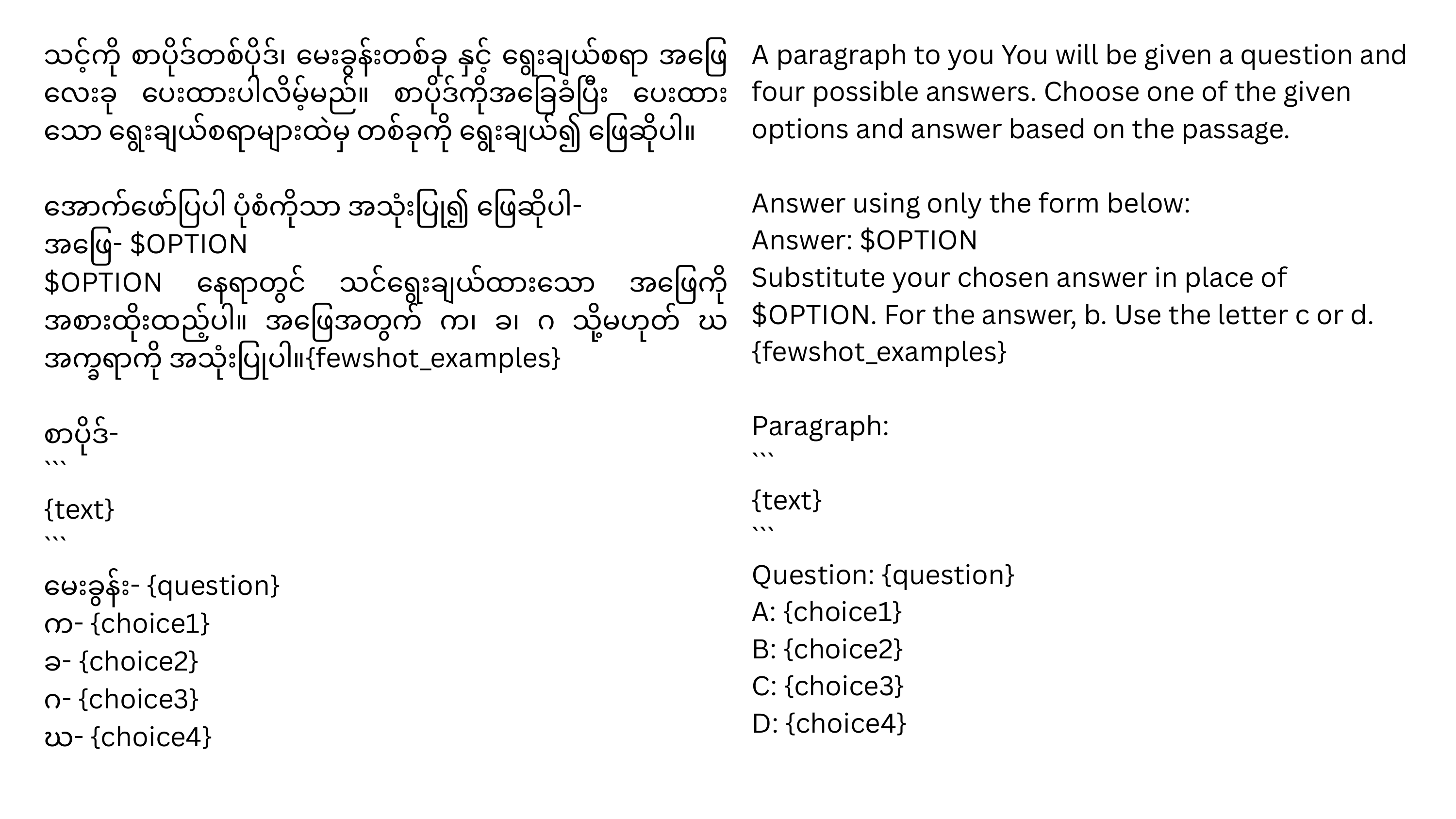}
    \subcaption{Question Answering}
\end{subfigure}\hfill
\begin{subfigure}[t]{0.48\textwidth}
    \centering
    \includegraphics[width=\linewidth]{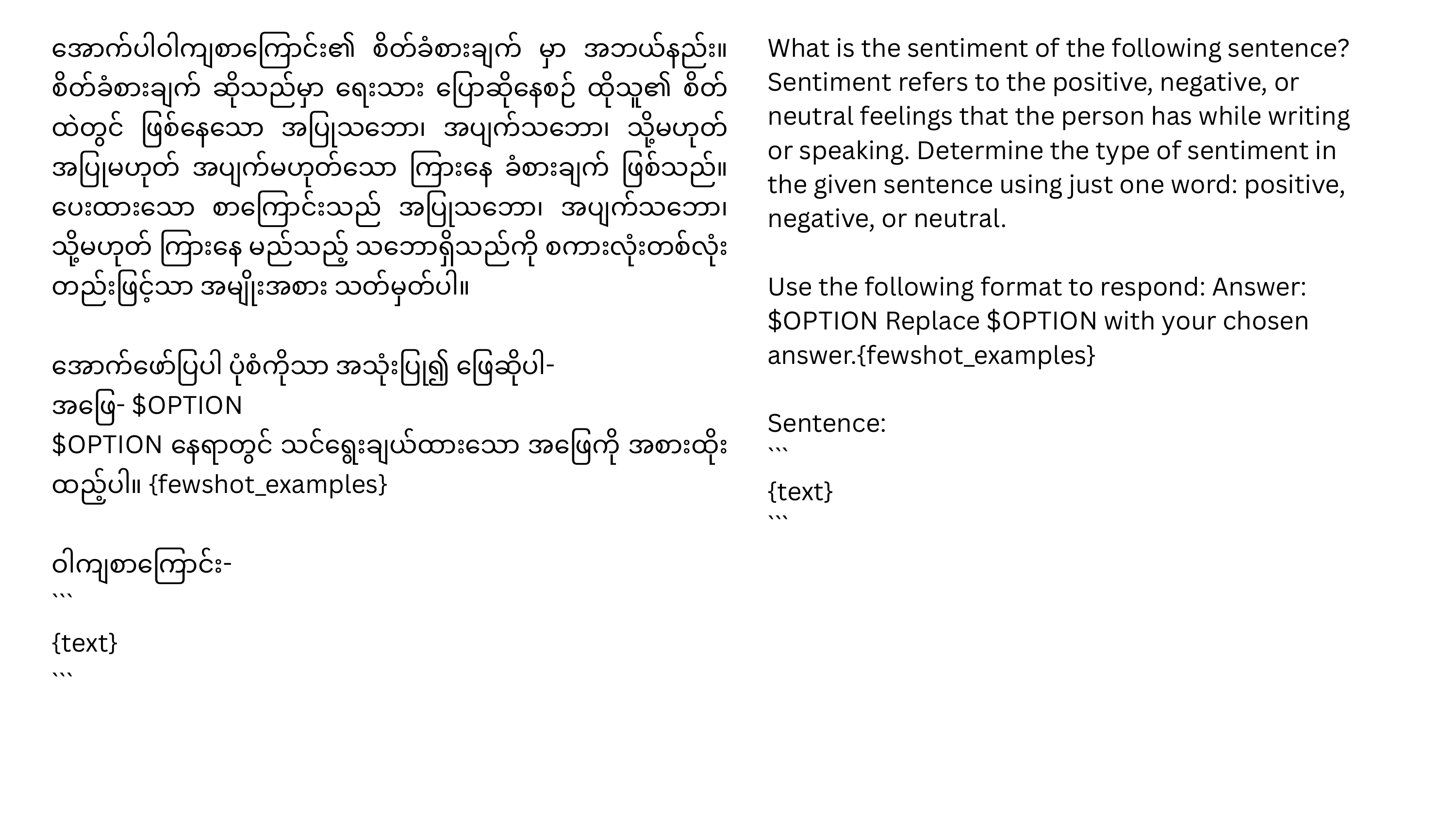}
    \subcaption{Sentiment Analysis}
\end{subfigure}

\vspace{0.5em}

\begin{subfigure}[t]{0.48\textwidth}
    \centering
    \includegraphics[width=\linewidth]{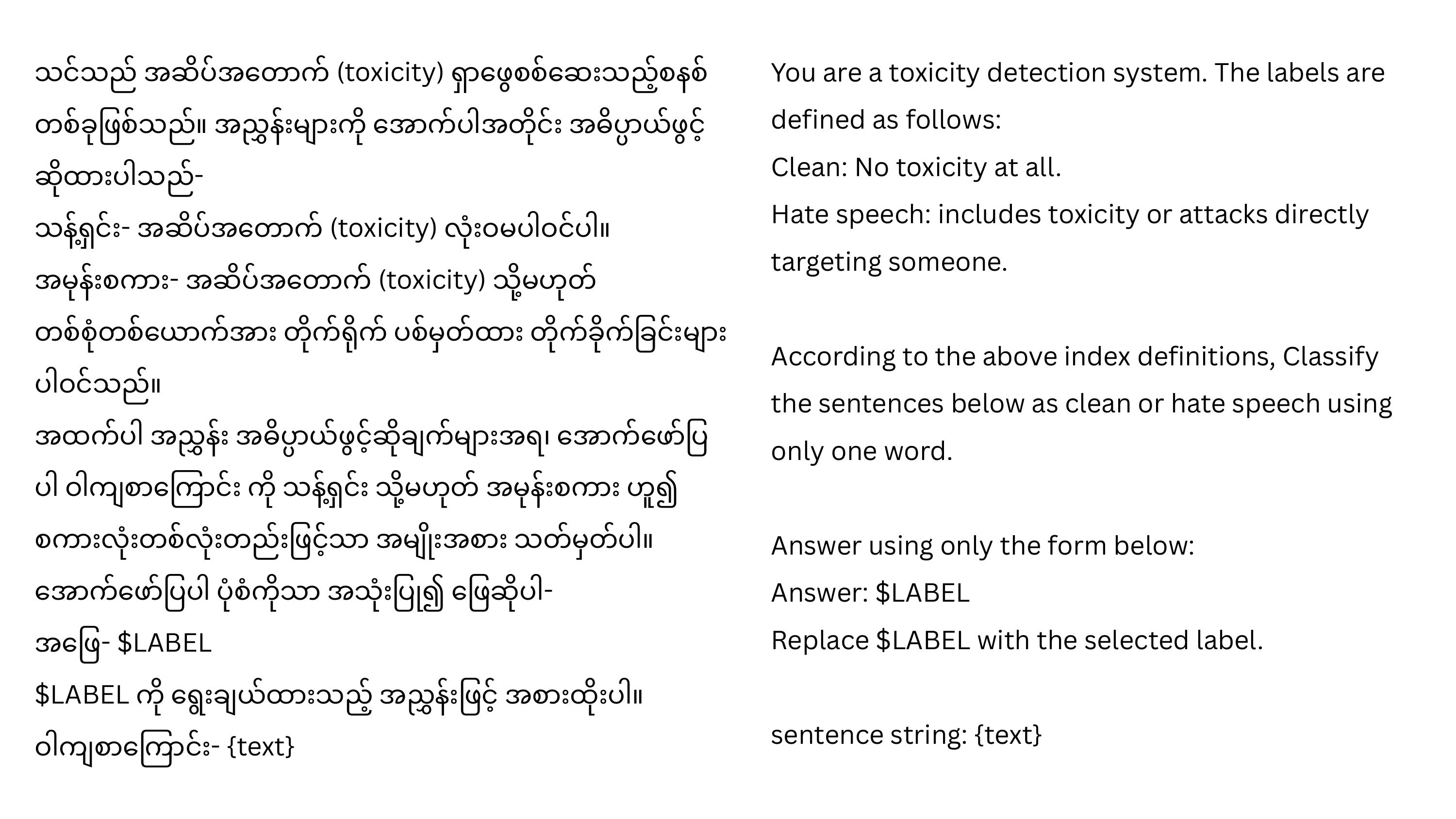}
    \subcaption{Toxicity Detection}
\end{subfigure}\hfill
\begin{subfigure}[t]{0.48\textwidth}
    \centering
    \includegraphics[width=\linewidth]{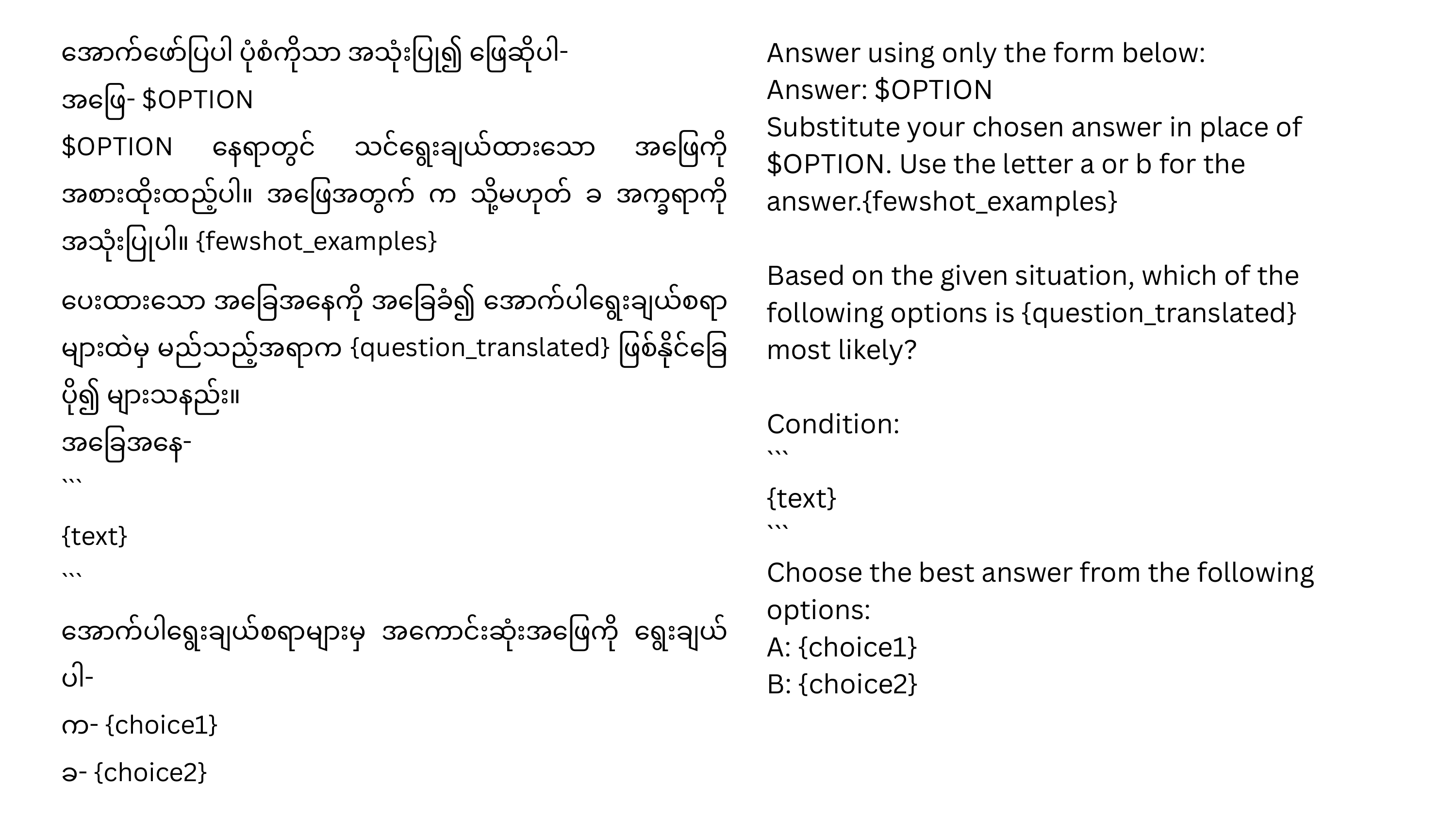}
    \subcaption{Causal Reasoning}
\end{subfigure}

\vspace{0.5em}

\begin{subfigure}[t]{0.48\textwidth}
    \centering
    \includegraphics[width=\linewidth]{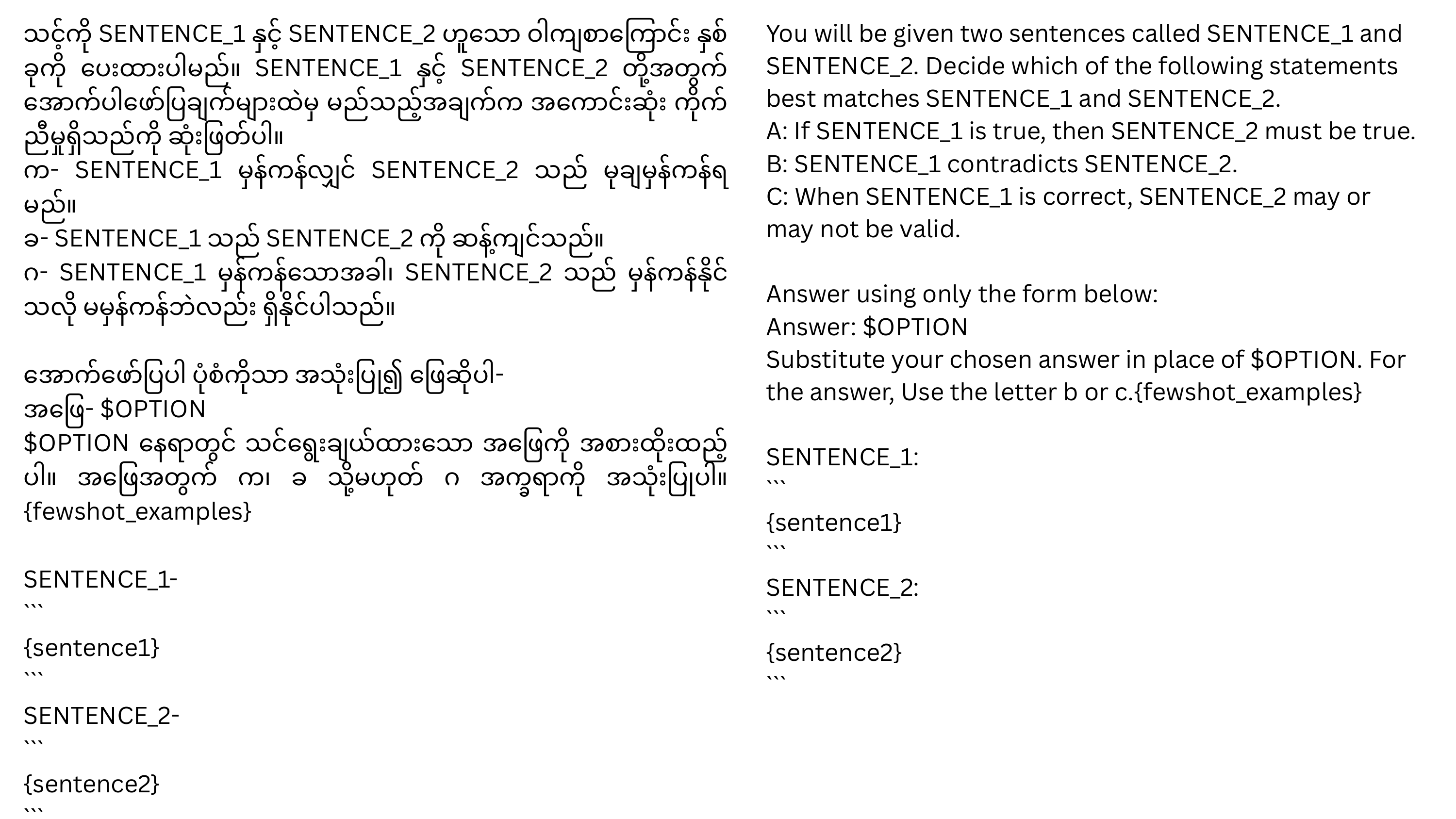}
    \subcaption{Natural Language Inference}
\end{subfigure}\hfill
\begin{subfigure}[t]{0.48\textwidth}
    \centering
    \includegraphics[width=\linewidth]{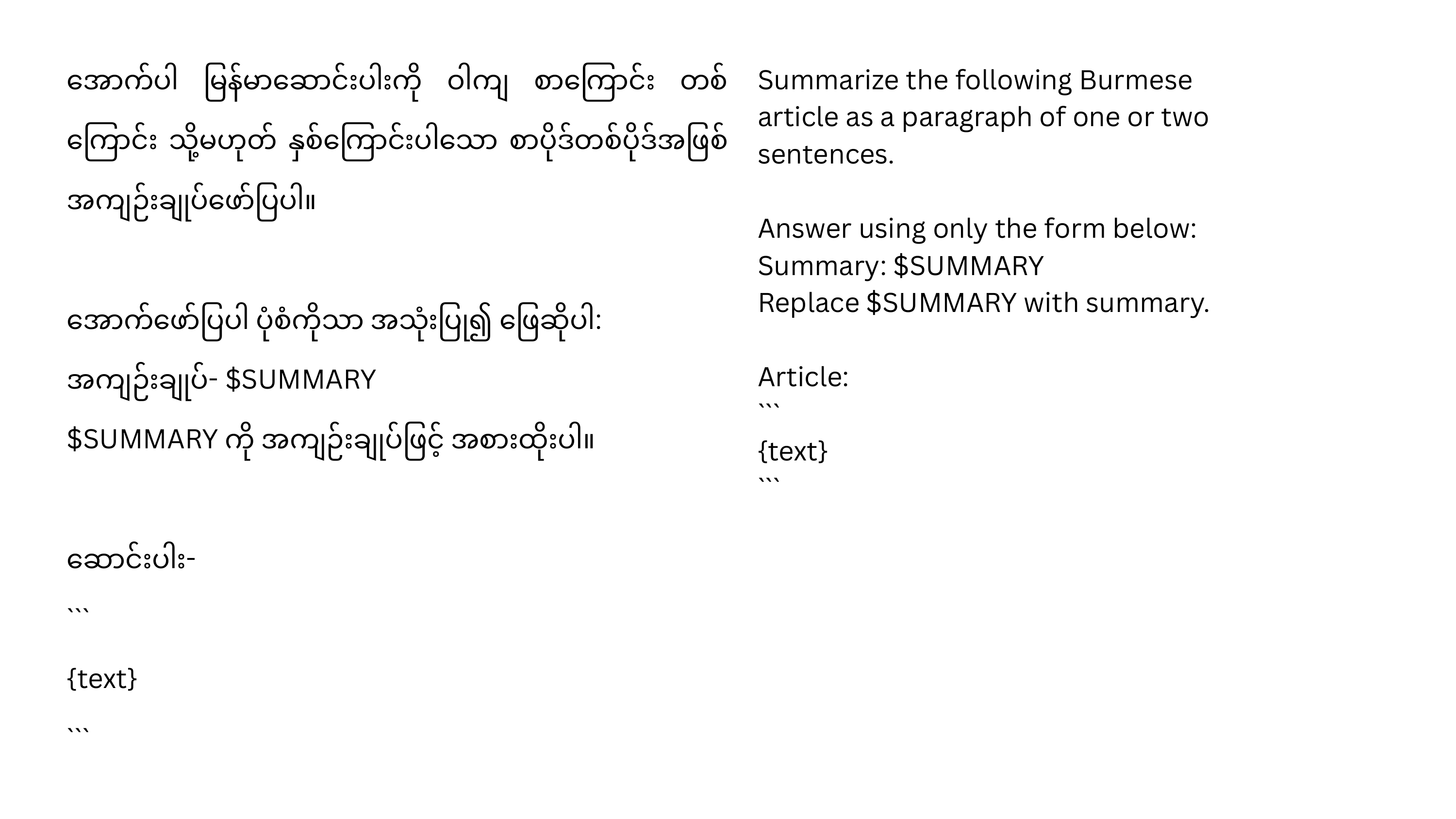}
    \subcaption{Abstractive Summarization}
\end{subfigure}

\vspace{0.5em}

\begin{subfigure}[t]{0.48\textwidth}
    \centering
    \includegraphics[width=\linewidth]{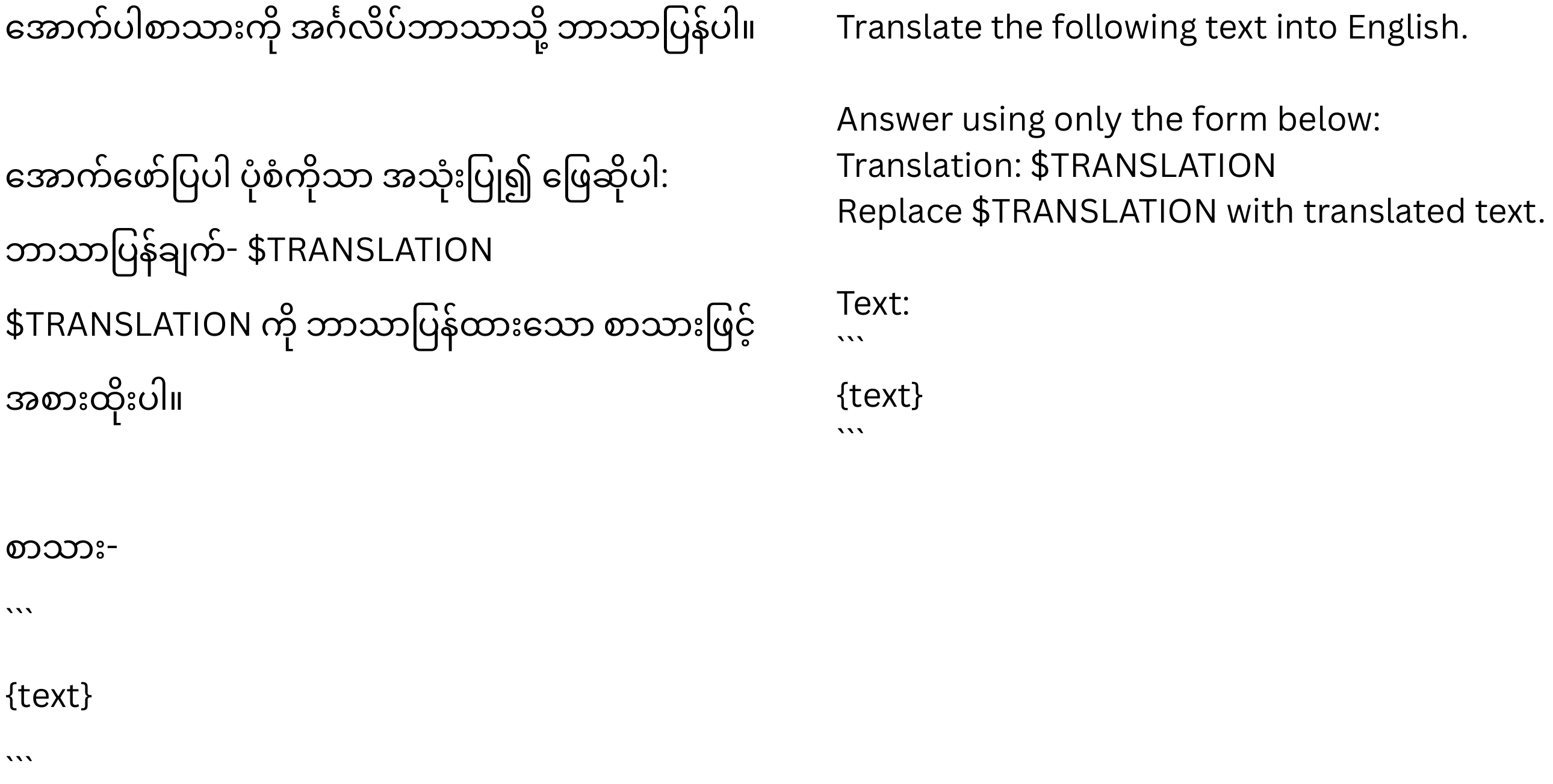}
    \subcaption{Machine Translation (to English)}
\end{subfigure}\hfill
\begin{subfigure}[t]{0.48\textwidth}
    \centering
    \includegraphics[width=\linewidth]{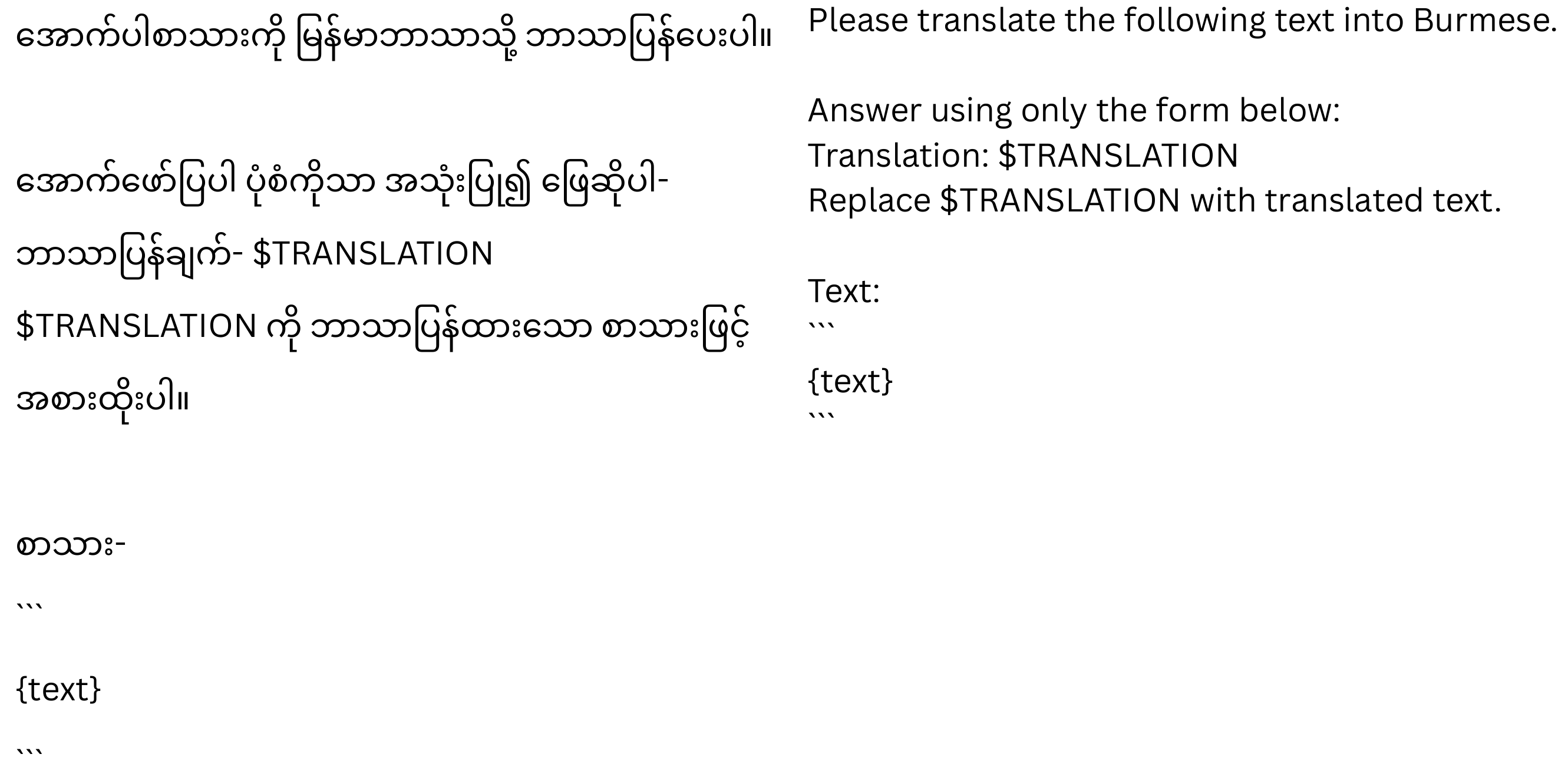}
    \subcaption{Machine Translation (to Myanmar)}
\end{subfigure}

\end{minipage}

\caption{Prompt templates used for \emph{BURMESE-SAN}. English prompt versions are also provided.}
\label{fig:prompt_templates}
\end{figure}

\newpage
\onecolumn

\renewcommand{\arraystretch}{1.2}
\begin{table}[H]
\centering
\tiny
\begin{tabular}{llr}
\toprule
\textbf{Models} & \textbf{Source Link} & \textbf{No. params} \\
\hline
\textit{Instruction Tuning Models} \\
\hline
ERNIE 4.5 & \url{https://huggingface.co/baidu/ERNIE-4.5-300B-A47B-PT} & 300B MoE \\
Qwen 3 A22B & \url{https://huggingface.co/Qwen/Qwen3-235B-A22B-Instruct-2507} & 235B MoE \\
Llama 4 Maverick & \url{https://huggingface.co/meta-llama/Llama-4-Maverick-17B-128E-Instruct} & 400B MoE \\
DeepSeek V3.1 & \url{https://huggingface.co/deepseek-ai/DeepSeek-V3.1} & 671B MoE \\
SEA-LION v4 (Qwen 3) & \url{https://huggingface.co/aisingapore/Qwen-SEA-LION-v4-32B-IT} & 32B \\
Gemma 3 VL & \url{https://huggingface.co/google/gemma-3-27b-it} & 27B \\
SEA-LION v4 (Gemma 3) & \url{https://huggingface.co/aisingapore/Gemma-SEA-LION-v4-27B-IT} & 27B \\
Llama 4 Scout & \url{https://huggingface.co/meta-llama/Llama-4-Scout-17B-16E-Instruct} & 109B MoE \\
Qwen 3 Next & \url{https://huggingface.co/Qwen/Qwen3-Next-80B-A3B-Instruct} & 80B MoE \\
Qwen 3 VL & \url{https://huggingface.co/Qwen/Qwen3-VL-32B-Instruct} & 32B \\
Kimi K2 Instruct 0905 & \url{https://huggingface.co/moonshotai/Kimi-K2-Instruct-0905} & 1040B MoE \\
Gemma 3 VL & \url{https://huggingface.co/google/gemma-3-12b-it} & 12B \\
Qwen 3 VL & \url{https://huggingface.co/Qwen/Qwen3-VL-32B-Instruct} & 32B \\
DeepSeek V3 & \url{https://huggingface.co/deepseek-ai/DeepSeek-V3-0324} & 671B MoE \\
SEA-LION v3 (Llama 3.1) & \url{https://huggingface.co/aisingapore/Llama-SEA-LION-v3-70B-IT} & 70B \\
Tulu 3 & \url{https://huggingface.co/allenai/Llama-3.1-Tulu-3-70B} & 70B \\
SEA-LION v4 (Qwen 3 VL) & \url{https://huggingface.co/aisingapore/Qwen-SEA-LION-v4-8B-VL} & 8B \\
Qwen 3 VL & \url{https://huggingface.co/Qwen/Qwen3-VL-8B-Instruct} & 8B \\
Qwen 2.5 & \url{https://huggingface.co/Qwen/Qwen2.5-72B-Instruct} & 72B \\
Qwen 2.5 & \url{https://huggingface.co/Qwen/Qwen2.5-32B-Instruct} & 32B \\
Qwen 3 VL & \url{https://huggingface.co/Qwen/Qwen3-VL-8B-Instruct} & 8B \\
Mistral Large 2411 & \url{https://huggingface.co/mistralai/Mistral-Large-Instruct-2411} & 123B \\
Qwen 3 A3B & \url{https://huggingface.co/Qwen/Qwen3-30B-A3B} & 30B MoE \\
Gemma 2 & \url{https://huggingface.co/google/gemma-2-27b-it} & 27B \\
SEA-LION v4 (Qwen 3 VL) & \url{https://huggingface.co/aisingapore/Qwen-SEA-LION-v4-4B-VL} & 4B \\
Llama 3.3 & \url{https://huggingface.co/meta-llama/Llama-3.3} & 70B \\
Llama 3.1 & \url{https://huggingface.co/meta-llama/Llama-3.1} & 70B \\
SEA-LION v3 (Llama 3.1) & \url{https://huggingface.co/aisingapore/Llama-SEA-LION-v3-8B-IT} & 8B \\
ERNIE 4.5 & \url{https://huggingface.co/baidu/ERNIE-4.5-21B-A3B-PT} & 21B MoE \\
Command A 03-2025 & \url{https://huggingface.co/CohereLabs/c4ai-command-a-03-2025} & 111B \\
SEA-LION v3 (Gemma 2) & \url{https://huggingface.co/aisingapore/Gemma-SEA-LION-v3-9B-IT} & 9B \\
Qwen 2.5 & \url{https://huggingface.co/Qwen/Qwen2.5-14B-Instruct} & 14B \\
Llama 3 & \url{https://huggingface.co/meta-llama/Meta-Llama-3-70B} & 70B \\
Apertus & \url{https://huggingface.co/swiss-ai/Apertus-70B-2509} & 70B \\
Sailor2 & \url{https://huggingface.co/sail/Sailor2-8B} & 8B \\
Tulu 3 & \url{https://huggingface.co/allenai/Llama-3.1-Tulu-3-8B-SFT} & 8B \\
MERaLiON 2 & \url{https://huggingface.co/MERaLiON/MERaLiON-2-10B} & 10B \\
Babel & \url{https://huggingface.co/Tower-Babel/Babel-83B} & 83B \\
Gemma 2 & \url{https://huggingface.co/google/gemma-2-9b-it} & 9B \\
Apertus & \url{https://huggingface.co/swiss-ai/Apertus-8B-2509} & 8B \\
Sailor2 & \url{https://huggingface.co/sail/Sailor2-20B} & 20B \\
Llama 3.1 & \url{https://huggingface.co/meta-llama/Llama-3.1-8B} & 8B \\
Qwen 2.5 & \url{https://huggingface.co/Qwen/Qwen2.5-7B-Instruct} & 7B \\
Babel & \url{https://huggingface.co/Tower-Babel/Babel-9B} & 9B \\
SeaLLMs V3 & \url{https://huggingface.co/SeaLLMs/SeaLLMs-v3-7B-Chat} & 7B \\
Command R+ 08-2024 & \url{https://huggingface.co/CohereLabs/c4ai-command-r-plus-08-2024} & 104B \\
phi-4 & \url{https://huggingface.co/microsoft/phi-4} & 14B \\
Aya Expanse & \url{https://huggingface.co/CohereLabs/aya-expanse-32b} & 32B \\
Command R 08-2024 & \url{https://huggingface.co/CohereLabs/c4ai-command-r-08-2024} & 32B \\
Olmo 2 0325 & \url{https://huggingface.co/allenai/OLMo-2-0325-32B-Instruct} & 32B \\
Ministral 2410 & \url{https://huggingface.co/mistralai/Ministral-8B-Instruct-2410} & 8B \\
Olmo 3 & \url{https://huggingface.co/allenai/Olmo-3-7B-Instruct} & 7B \\
Llama 3 & \url{https://huggingface.co/meta-llama/Meta-Llama-3-8B} & 8B \\
Command R7B 12-2024 & \url{https://huggingface.co/CohereLabs/c4ai-command-r7b-12-2024} & 7B \\
Aya Expanse & \url{https://huggingface.co/CohereLabs/aya-expanse-8b} & 8B \\
Mistral Small 3.1 2503 & \url{https://huggingface.co/mistralai/Mistral-Small-3.1-24B-Instruct-2503} & 24B \\
Olmo 2 1124 & \url{https://huggingface.co/allenai/OLMo-2-1124-7B-Instruct} & 7B \\
Olmo 2 1124 & \url{https://huggingface.co/allenai/OLMo-2-1124-13B-Instruct} & 13B \\
SEA-LION v4 (Gemma 3 VL) & \url{https://huggingface.co/aisingapore/Gemma-SEA-LION-v4-4B-VL} & 4B \\
Gemma 3 VL & \url{https://huggingface.co/google/gemma-3-4b-it} & 4B \\
Qwen 3 & \url{https://huggingface.co/Qwen/Qwen3-4B} & 4B \\
Qwen 3 VL & \url{https://huggingface.co/Qwen/Qwen3-VL-4B-Instruct} & 4B \\
SEA-LION v4 (Apertus) & \url{https://huggingface.co/aisingapore/Apertus-SEA-LION-v4-8B-IT} & 8B \\
\hline
\textit{Reasoning Models} \\
\hline
Qwen 3 (Thinking) & \url{https://huggingface.co/Qwen/Qwen3-235B-A22B-Thinking-2507} & 235B MoE \\
DeepSeek V3.1 Thinking & \url{https://huggingface.co/deepseek-ai/DeepSeek-V3.1} & 671B MoE \\
Qwen 3 Next (Thinking) & \url{https://huggingface.co/Qwen/Qwen3-Next-80B-A3B-Thinking} & 80B MoE \\
Deepseek R1 0528 & \url{https://huggingface.co/deepseek-ai/DeepSeek-R1-0528} & 671B MoE \\
Qwen 3 (Thinking) & \url{https://huggingface.co/Qwen/Qwen3-30B-A3B-Thinking-2507} & 30B MoE \\
Qwen 3 (Thinking) & \url{https://huggingface.co/Qwen/Qwen3-32B} & 32B \\
SEA-LION v3.5 R (Llama) & \url{https://huggingface.co/aisingapore/Llama-SEA-LION-v3.5-70B-R} & 70B \\
Qwen 3 (Thinking) & \url{https://huggingface.co/Qwen/Qwen3-14B} & 14B \\
Qwen 3 (Thinking) & \url{https://huggingface.co/Qwen/Qwen3-8B} & 8B \\
GPT OSS & \url{https://huggingface.co/openai/gpt-oss-20b} & 20B MoE mxfp4 \\
Qwen 3 (Thinking) & \url{https://huggingface.co/Qwen/Qwen3-VL-32B-Thinking} & 4B \\
QwQ & \url{https://huggingface.co/Qwen/QwQ-32B-Preview} & 32B \\
Reka Flash 3.1 & \url{https://huggingface.co/RekaAI/reka-flash-3.1} & 21B \\
SEA-LION v3.5 R (Llama) & \url{https://huggingface.co/aisingapore/Llama-SEA-LION-v3.5-8B-R} & 8B \\
Olmo 3 Think & \url{https://huggingface.co/allenai/Olmo-3-32B-Think} & 32B \\
Olmo 3 Think & \url{https://huggingface.co/allenai/Olmo-3-7B-Think} & 7B \\
\hline
\end{tabular}
\caption{Links and sizes of models evaluated in our current experiments.}
\label{tab:model_links_sizes_current}
\end{table}

\end{document}